\documentclass[sigconf,screen]{acmart}

\usepackage{multirow}
\usepackage{subcaption}

\usepackage{hyperref}
\hypersetup{
    colorlinks=true,
    linkcolor=NavyBlue,
    citecolor=NavyBlue,
    urlcolor=NavyBlue
}

\AtBeginDocument{%
  \providecommand\BibTeX{{%
    \normalfont B\kern-0.5em{\scshape i\kern-0.25em b}\kern-0.8em\TeX}}}

\setcopyright{acmlicensed}
\copyrightyear{2024}
\acmYear{2024}
\acmDOI{XXXXXXX.XXXXXXX}

\acmConference[Preprint]{Preprint}{February 12, 2024}{}
\acmISBN{XXX-X-XXXX-XXXX-X/XX/XX}

\begin{document}

\title[Beyond the Mud]{Beyond the Mud: Datasets and Benchmarks for Computer Vision in Off-Road Racing}





\author{Jacob Tyo}
\email{jtyo@cs.cmu.edu}
\affiliation{%
  \institution{DEVCOM Army Research Laboratory}
  \institution{Carnegie Mellon University}
  \streetaddress{5000 Forbes Ave}
  \city{Pittsburgh}
  \state{Pennsylvania}
  \country{USA}
  \postcode{15239}
}

\author{Motolani Olarinre}
\email{tolani@cs.cmu.edu}
\affiliation{%
  \institution{Carnegie Mellon University}
  \streetaddress{5000 Forbes Ave}
  \city{Pittsburgh}
  \state{Pennsylvania}
  \country{USA}
  \postcode{15239}
}

\author{Youngseog Chung}
\email{young@cs.cmu.edu}
\affiliation{%
  \institution{Carnegie Mellon University}
  \streetaddress{5000 Forbes Ave}
  \city{Pittsburgh}
  \state{Pennsylvania}
  \country{USA}
  \postcode{15239}
}

\author{Zachary C. Lipton}
\email{zlipton@cs.cmu.edu}
\affiliation{%
  \institution{Carnegie Mellon University}
  \streetaddress{5000 Forbes Ave}
  \city{Pittsburgh}
  \state{Pennsylvania}
  \country{USA}
  \postcode{15239}
}

\renewcommand{\shortauthors}{Tyo, et al.}

\begin{abstract}
    Despite significant progress in optical character recognition (OCR) 
    and computer vision systems, 
    robustly recognizing text 
    and identifying people in images taken 
    in unconstrained \emph{in-the-wild} 
    environments remain an ongoing challenge.
    However, 
    such obstacles must be overcome 
    in practical applications of vision systems, 
    such as identifying racers in photos 
    taken during off-road racing events. 
    To this end, 
    we introduce two new challenging real-world datasets - 
    the off-road motorcycle Racer Number Dataset (RND) 
    and the Muddy Racer re-iDentification Dataset (MUDD) - 
    to highlight the shortcomings of current methods
    and drive advances in OCR and person re-identification (ReID) under extreme conditions. 
    These two datasets feature over 6,300 images 
    taken during off-road competitions 
    which exhibit a variety of factors 
    that undermine even modern vision systems, 
    namely mud, complex poses, and motion blur.
    We establish benchmark performance on both datasets 
    using state-of-the-art models. 
    Off-the-shelf models transfer poorly, 
    reaching only 15\% end-to-end (E2E) F1 score on text spotting, 
    and 33\% rank-1 accuracy on ReID. 
    Fine-tuning yields major improvements, 
    bringing model performance to 53\% F1 score for E2E text spotting
    and 79\% rank-1 accuracy on ReID, 
    but still falls short of good performance. 
    Our analysis exposes open problems in real-world OCR 
    and ReID that necessitates domain-targeted techniques. 
    With these datasets and analysis of model limitations, 
    we aim to foster innovations in handling real-world conditions 
    like mud and complex poses 
    to drive progress in robust computer vision.
    All data was sourced from \url{PerformancePhoto.co}, 
    a website used by professional motorsports photographers, 
    racers, and fans. 
    The top-performing text spotting 
    and ReID models are deployed on this platform 
    to power real-time race photo search. 
    All code and data 
    can be found at \url{https://drive.google.com/file/d/1YFlQLh3rorLLvG8reYxHiqs7HkWf1lJF/view?usp=sharing}.
\end{abstract}




\begin{teaserfigure}
    \centering
    \begin{subfigure}[b]{0.54\textwidth}
        \centering
        \includegraphics[width=\textwidth]{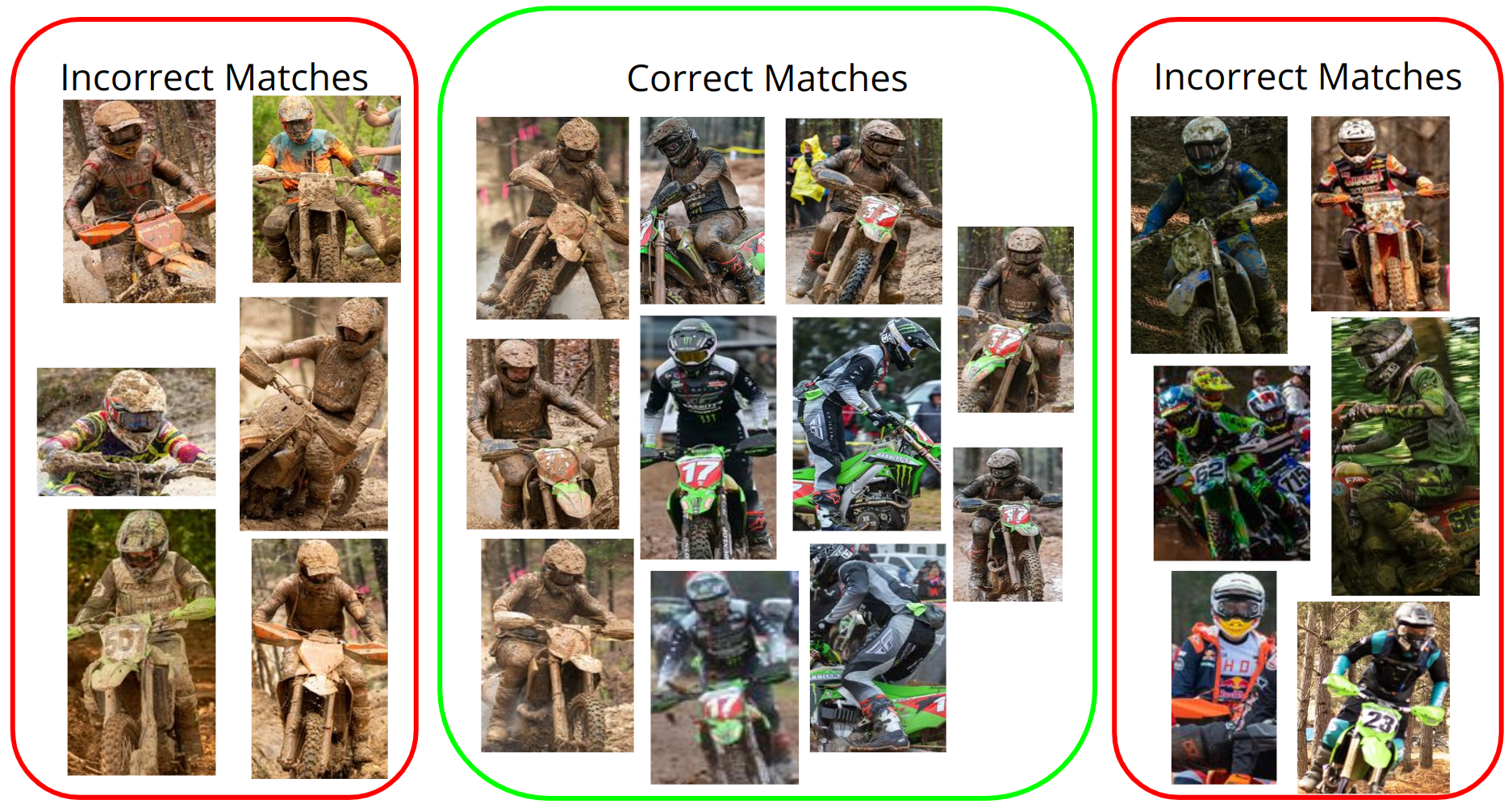}
        \caption{}
    \end{subfigure}
    \hfill 
    \begin{subfigure}[b]{0.45\textwidth}
        \centering
        \includegraphics[width=\textwidth]{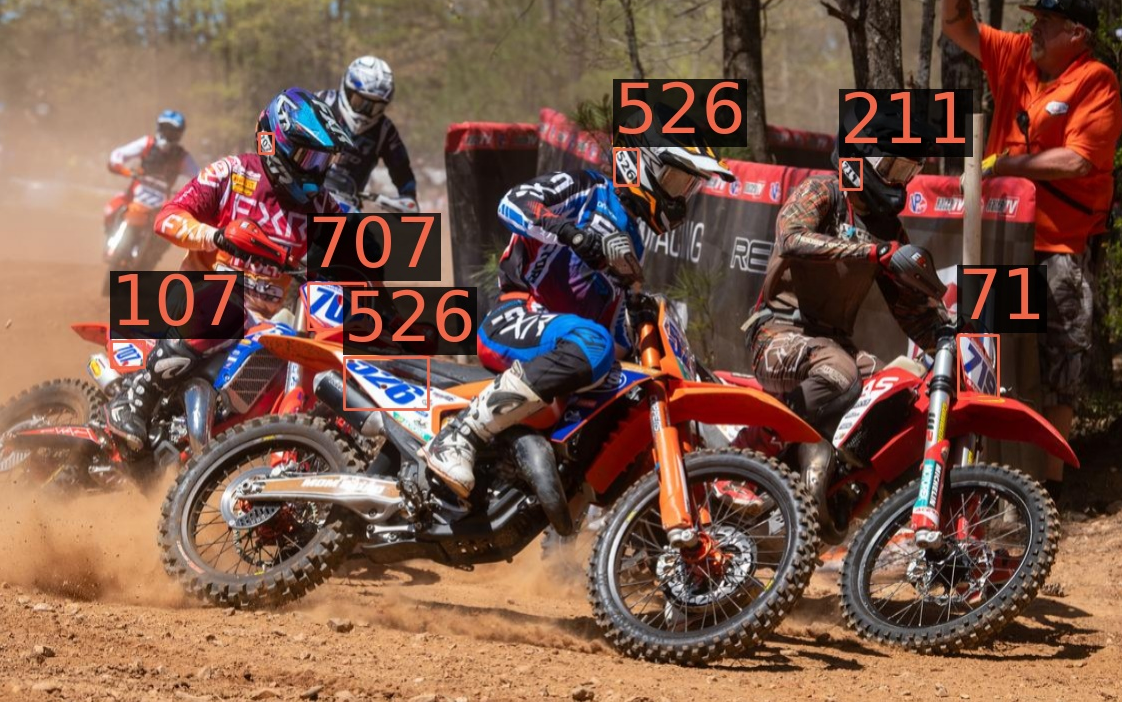}
        \caption{}
    \end{subfigure}
    \caption{(a) Example of muddy images, grouped by the person re-identification label, emphasizing both the text spotting and person re-identification difficulty. (b) A clean (i.e. not muddy) example from the text spotting dataset. 
    }
    \label{fig:headline}
\end{teaserfigure}


\maketitle

\section{Introduction}

Robustly recognizing visual concepts like text and people in unconstrained real-world environments remains an ongoing challenge in computer vision. 
While the fields of document text recognition and detection (jointly referred to as \emph{text spotting}) as well as person re-identification (ReID) have made significant advances in recent years owing to deep learning methods and large annotated datasets, performance still tends to degrade under difficult lighting, occlusion, unusual perspectives, motion blur, and other factors not well-represented in training data~\citep{Appalaraju_2021_ICCV, shashirangana2020automated, netzer2011reading, ye2021deep, gou2018systematic, textinthewild}.

For example, state-of-the-art optical character recognition (OCR) models can now transcribe scanned documents and signage with over 99\% word accuracy~\citep{fujitake2024dtrocr}. 
However, accurately reading text on objects in cluttered scenes or under heavy occlusion is still limited to around 80\%~\citep{zhao2023clip4str, Ye_2023_CVPR} in benchmarks like Total-Text~\citep{ch2017total} and COCO-Text~\citep{veit2016coco}. 
Similarly, person re-identification models can reliably match people across different camera views when imaging conditions are consistent. 
But model accuracy drops substantially in the presence of occlusions or rarely seen poses~\citep{ye2021deep}.

In this work, we focus on one particularly challenging real-world domain that exposes the brittleness of current techniques - off-road motorcycle racing. 
Correctly identifying riders and detecting racing numbers is not only important for accurate photo searching, but also for sports analytics, automatic timing and scoring, photo searching, and more. 
However, this setting induces several factors that significantly undermine the capability of existing vision models:
\begin{enumerate}
    \item Heavy mud spatters accumulate and obscure text 
    and riders in irregular, unprecedented ways. 
    Mud caking exhibits occlusion patterns not seen 
    in typical datasets.
    \item Riders strike complex poses including jumps, crashes, and wheelies, 
    among others which are rarely depicted under standard walking scenarios. 
    These confuse models tuned to pedestrian datasets.
    \item Sizes of number tags vary from a few inches to over a foot tall, 
    placed in diverse layouts on bikes and rider gear, 
    and in an ever-increasing number of fonts. 
    \item Racing conditions cause glare, blurring, 
    awkward angles, and low-resolution imagery 
    not prevalent in controlled datasets.
    As a race progresses, 
    the appearance of a single racer can change dramatically.
\end{enumerate}

To spur progress in this domain, 
we introduce two real-world datasets 
captured from off-road competitions. 
The \emph{off-road Racer Number Dataset} (RND) 
contains 2,411 images exhibiting riders engaged in races. 
Each image is annotated with bounding boxes 
around visible racer numbers on bikes, jerseys, or helmets, 
along with the number text transcription (e.g. ``15A"). 
In total, there are over 5,500 labeled rider numbers 
spanning challenges like mud, awkward perspectives, low resolution, and more.

The second dataset focuses specifically 
on the rider re-identification task. 
The \emph{MUddy Racer re-iD Dataset} (MUDD) 
includes 3,906 images capturing 150 riders 
across 10 different off-road events. 
Each rider crop is annotated with a unique identity label. 
The imagery is subject to real-world variations in mud, 
lighting, pose, blurring, age, outfit changes, 
crashes, and more over each race.

We benchmark state-of-the-art vision models on both datasets and find that pretrained models transfer poorly, 
with accuracies under 30\% in both text spotting and ReID. 
Even after fine-tuning, 
the best models 
achieve only 53\% F1 score for end-to-end racer number recognition, 
and just 79\% rank-1 accuracy on rider matching. 
Our analysis exposes open problems in handling mud occlusion, 
appearance change, layouts, resolution, and other factors. 
Through domain-targeted data, benchmarks, and analysis, 
we aim to spur innovation advancing robust capabilities.

Our contributions are:
\begin{itemize}
    \item RnD: an off-road motorcycle Racer number Dataset 
        containing 2,411 images with 5,578 labeled numbers sampled from
        professional photographers at 50 distinct off-road races. 
    \item MUDD: a Muddy racer re-identification Dataset 
        for person re-id, containing 3,906 images of 150 identities captured over
        10 off-road events by 16 professional motorsports photographers. 
    \item Initial benchmarking of state-of-the-art models, which reveal
        limitations on these datasets and show substantial room for
        further improvement.
    \item Analysis of failure cases that provide insights to guide
        future research on robust re-identification and text spotting
        for sports analytics and computer vision broadly.
\end{itemize}

\begin{figure*}[h]
    \centering
    \begin{subfigure}[b]{0.1335\textwidth}
        \centering
        \includegraphics[width=\textwidth]{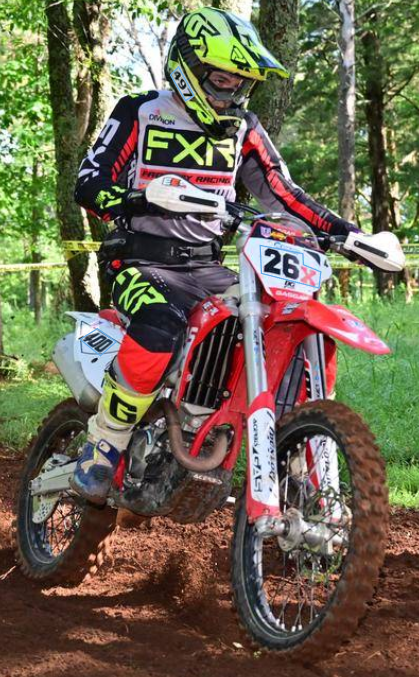}
        \caption{}
    \end{subfigure}
    \hfill
    \begin{subfigure}[b]{0.1325\textwidth}
        \centering
        \includegraphics[width=\textwidth]{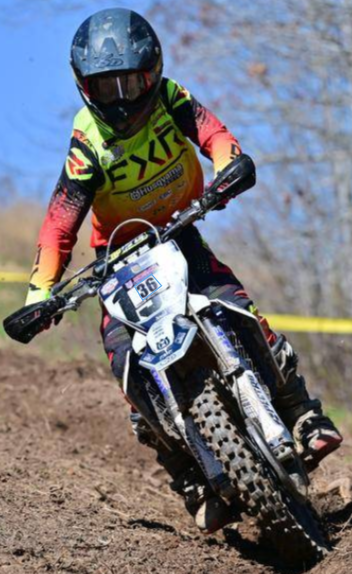}
        \caption{}
        \label{img:overlap_glare}
    \end{subfigure}
    \hfill 
    \begin{subfigure}[b]{0.226\textwidth}
        \centering
        \includegraphics[width=\textwidth]{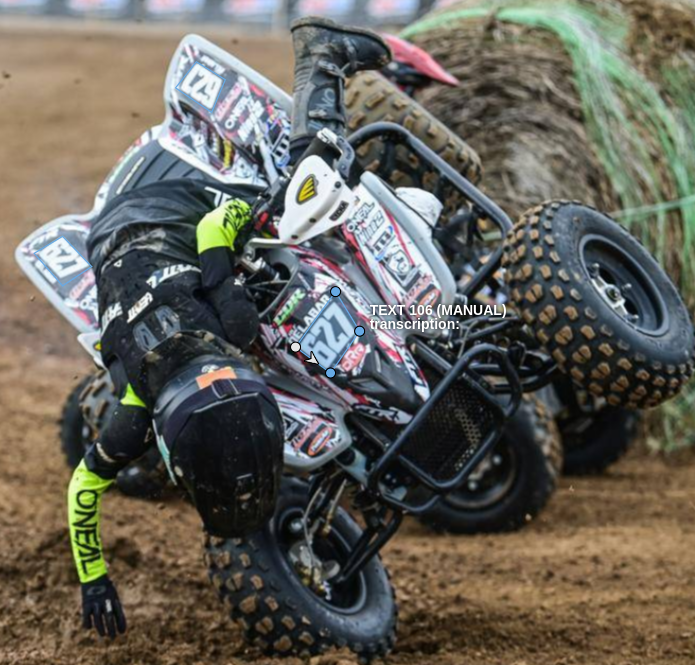}
        \caption{}
        \label{img:atv_crash}
    \end{subfigure}
    \hfill
    \begin{subfigure}[b]{0.1585\textwidth}
        \centering
        \includegraphics[width=\textwidth]{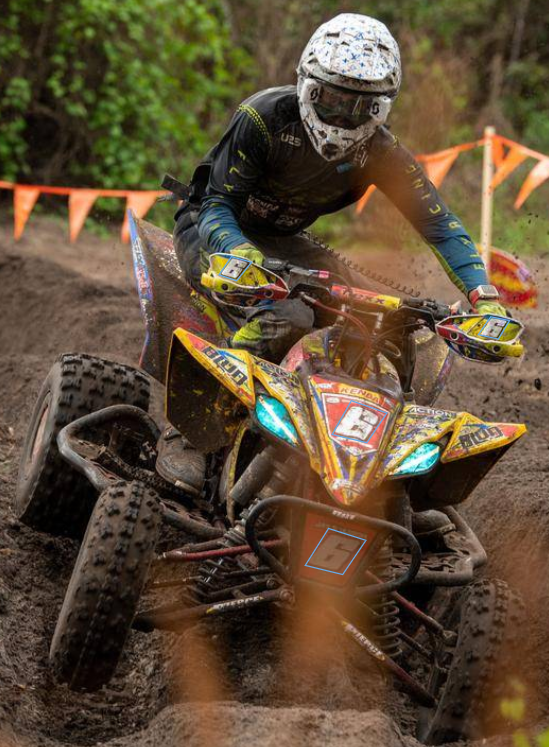}
        \caption{}
    \end{subfigure}
    \hfill
    \begin{subfigure}[b]{0.116\textwidth}
        \centering
        \includegraphics[width=\textwidth]{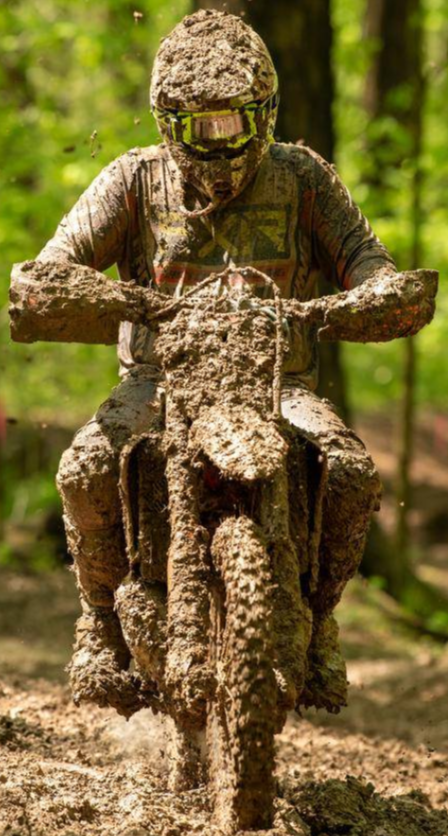}
        \caption{}
        \label{img:completemud}
    \end{subfigure}
    \hfill
    \begin{subfigure}[b]{0.21\textwidth}
        \centering
        \includegraphics[width=\textwidth]{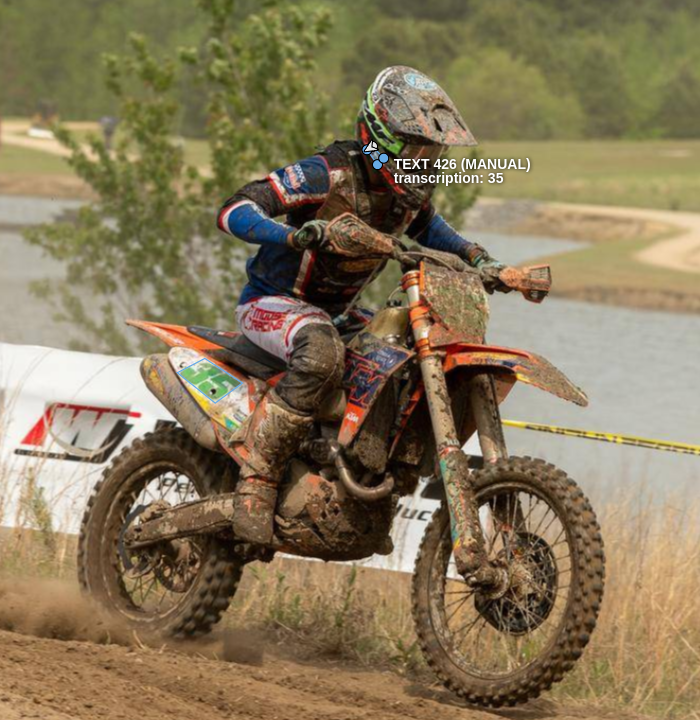}
        \caption{}
    \end{subfigure}
    \caption{Beyond the Mud Dataset Examples:
    (a) Racers can have multiple, non-matching, numbers,
    (b) glare renders some numbers impossible to read, 
    (c) a crashing racer,
    (d) occlusions from vegetation,
    (e) an extreme example of mud, 
    (f) a stereotypical amount of mud.}
    \label{fig:how_labeled}
\end{figure*}

\section{The Datasets}

We introduce two large-scale datasets
constructed from real-world off-road motorcycle competition imagery 
from \url{PerformancePhoto.co}. 
The dataset was collected from over 
50 different off-road motorcycle and ATV races, 
from 16 different professional photographers, 
using a wide range of high-end cameras. 
This includes both professional and amateur races 
across a wide range of geographical locations in the United States.
All labeling was done manually by three human annotators. 
Both datasets are separated into 72\%/8\%/20\% train/validation/test sets.

The data exhibits extreme diversity in uncontrolled conditions, 
including: 
\begin{itemize}
    \item Heavy mud occlusion - 
    Racers accumulate significant mud spatters 
    and caking throughout the duration of each race.
    This represents a unique occlusion pattern not present 
    in existing re-id datasets.
    \item Complex poses---Racers exhibit varied poses including leaning, 
    jumping, crashes, and more that are uncommon during regular walking.
    \item Distance and resolution---Images captured from a distance with small, 
    low-resolution racer crops.
    \item Dynamic lighting---Outdoor conditions cause glare, 
    shadows, and varying levels of exposure.
    \item Clothing---Jerseys and numbers that could ease 
    re-id are often obscured by mud, gear, and positioning.
    \item Motion blur---Racers maneuver at high speeds causing 
    motion-blurring effects.
\end{itemize}
Given the challenges associated with this dataset, 
one may perceive the proper re-identification of racers 
to be nearly impossible even for humans, 
less for any vision system.
However, there are several consistent features even among the most extreme mud. 
Some of those features include the helmet shape,  
hydration system shape, the shape of the handguards, 
boot shape, and other similar characteristics.

\subsection{RnD Dataset}

The off-road motorcycle Racer number Dataset (RnD)
is comprised of 2,411
images.
Each image depicts motorcycle racers engaged in competitive events,
with visible racer numbers on themselves and their motorcycles, 
labeled with bounding box annotations and text transcriptions.

Racers can have anywhere from one to as many as 20 numbers
located on their body and motorcycle.
The common locations for a number include
the front and sides of the motorcycle,
on the cheeks of the racer's helmet,
and on the back of the racers jersey.
However, in rare cases,
numbers can also be seen on the wheels
and handguards.
The numbers on a single racer and vehicle
do not need to all be the same number.
Commonly,
the numbers on the helmet do not match the numbers on
the motorcycle,
and the number on the front of the motorcycle
does not need to match the number on the side.
It is also common for numbers to only be present on the racer,
but not on the motorcycle.
Figure~\ref{fig:how_labeled} highlights some of these examples.

In RnD, there is a total of 5,578 racer number annotations.
The numbers can span from 1 to 5 characters in length,
optionally including alphabetical characters
(e.g., adding a letter to the end of a number is
a common modifier -
for convenience, we still refer to all of these as \emph{numbers}).
6\% of the dataset includes numbers that have
alphabetical characters in addition to the numerics.
The train/validation/test split was done randomly.
Only the racer numbers were annotated
(instead of all visible text) by one of the authors.
All visible racer numbers were tightly bound by a polygon
(i.e. the bounding box),
and each polygon is tagged with the characters contained within 
(i.e. the number).
If a character was ambiguous or unclear, 
it was labeled with a ``\#'' symbol. 
Only 
the humanly identifiable text was transcribed. 
Racer numbers that were fully occluded 
or too blurry to discern were not annotated.

The annotation task was restricted 
to only use the context of each 
individual bounded region. 
The full image context could not be used 
to infer ambiguous numbers based on 
other instances of that racer's number 
elsewhere on the motorcycle. 
This stimulates the local context available 
to optical character recognition models.
Figure~\ref{fig:how_labeled} depicts a few characteristic images of this dataset. 
Glare (8\%), shadows (7\%), blur (3\%), dust (2\%), and mud (44\%) 
make for a very challenging task. 
Several other factors even further complicate the text-spotting task, such as 
numbers placed on top of other numbers
and odd orientations from crashes. 

\subsection{MUDD: Muddy Racer Re-Identification Dataset}

MUDD contains 3906 images 
capturing 150 identities
across 10 different off-road events from 
the Grand National Cross Country (GNCC) racing series. 
We used YOLOX~\citep{ge2021yolox} 
to detect the bounding boxes for people, 
which were then manually labeled by our annotators. 
Importantly,
there is no identity overlap between the train/validation/test sets. 
Furthermore, 
both the validation and test sets are further broken into a \emph{query} 
and \emph{gallery} set. 
Based on the query set, 
the gallery images are ranked based on similarity for evaluation. 
Data was only used from 10 events because 
we only matched the identities of racers within events. 
It is more difficult to tell apart two people from a single race 
than two people from a different race, 
solely because of geographical hints in the background. 

In labeling this re-identification dataset, 
we created an initial clustering using off-the-shelf 
re-id~\citep{zhou2019omni} 
and text-spotting models~\citep{krylov2021open}. 
First by detecting numbers and people, 
then for each overlapping person and number, 
ranking every other person based on their similarity. 
Our human annotators then manually review and create ground truth clusters, 
aided by this initial ranking. 
This is discussed in more detail in Appendix~\ref{sec:mudd-labeling-info}. 
Just as with RnD, 
this dataset contains all of the difficulties of off-road racing
like mud, dust, glare, blur, etc.

\section{Experiments}

We conducted experiments 
to benchmark the performance of both state-of-the-art off-the-shelf 
methods, as well as their fine-tuned counterparts on 
both person search (i.e. the re-id task) and text spotting in this motorcycle racing setting.
We also leverage the attributes of each image to analyze
how mud, glare, shadow, dust, and blur affect performance. 
All experimentation was done on four NVIDIA Tesla V100 GPUs. 
Hyperparameter searching was performed for each experiment
using the training and validation sets. 
Then the best-performing hyperparameters were used to train a final model 
on the combined training and validation set, 
for evaluation on the test set. 
Further details on the exact configuration of the hyperparameter search can be 
found in Appendix~\ref{}. 

\subsection{Spotting Racer Numbers}

Our experiments leverage two state-of-the-art scene text spotting models: 
\begin{itemize}
    \item \textbf{YAMTS}: Yet Another Mask Text Spotter \cite{krylov2021open}

     YAMTS is a Mask R-CNN-based model with an additional recognition head
    for end-to-end scene text spotting.
    A ResNet-50~\cite{he2016deep} is used for text detection, 
    with a convolutional text encoder and a GRU decoder. 
    \item \textbf{SwinTS}: Swin Text Spotter \cite{huang2022swintextspotter}
    
    The Swin Text Spotter is an end-to-end 
    Transformer-based model that improves detection 
    and recognition synergy through a recognition conversion module.
    A feature pyramid network is used to decrease the sensitivity to text size, 
    and the recognition conversion model 
    enables joint optimization of the detection 
    and recognition losses. 
\end{itemize}

For both models, 
we first benchmark their performance on the RnD test set using 
their published pre-trained weights. 
Afterwards, we fine-tune these models further
on the RnD training set and evaluate their performance again.
The hyperparameter search, 
detailed in Appendix~\ref{sec:hyperparameters},
revealed the best parameters to be a cosine annealing learning rate
schedule with a warm-up, 
using a total batch size of 32, 
with the random scaling and rotation data augmentations. 
The learning rate starts at 1e-6
and is then raised to 1e-3 after 1,000 iterations, 
and then annealed back down to 1e-6 over the remainder of training.
These hyperparameters were used to fine-tune the models 
over 150 epochs.
The fine-tuned models are evaluated on the RnD test set.

\subsubsection{Text Spotting Evaluation Metrics}

Following the standard evaluation protocol~\cite{huang2022swintextspotter, Ye_2023_CVPR}, 
we report results for both the text detection 
and end-to-end recognition tasks.
For detection, we compute precision, recall, and F1-score, which we denote \textit{Det-P}, \textit{Det-R}, and \textit{Det-F1} respectively. 
A predicted box was considered a true positive 
if it overlapped with a ground truth box 
by at least 50\% intersection over the union.
For end-to-end recognition, 
we report precision, recall, and F1-score at the sequence level, and we likewise denote these metrics as \textit{E2E-P}, \textit{E2E-R}, and \textit{E2E-F1}. 
A predicted text sequence was considered correct 
only if it exactly matched the ground truth transcription for the corresponding ground truth box.

\subsection{Re-Identifying Racers}

We evaluated the performance of two models on MUDD in three settings:
\begin{itemize}
    \item \textbf{Off-the-shelf}: Pre-trained state-of-the-art re-id models applied directly to MUDD. 
    \item \textbf{Random Initialization}: Models trained from random initialization only on MUDD. 
    \item \textbf{Transfer}: Person re-identification pre-trained models 
    fine-tuned on MUDD. 
\end{itemize}
The models used were strong open-source implementations of CNN-based architectures, specifically OSNet~\citep{zhou2019omni} and ResNet50~\citep{he2016deep}.
The hyperparameter search detailed in Appendix~\ref{sec:hyperparameters} revealed the best parameters to be the Adam optimizer using the triplet loss, with data augmentation of random flips, color jitter, and random crops. 
A cosine learning rate schedule was used with a maximum learning rate of $0.0003$, over 100 epochs. 
The mean and standard deviation for each experiment are reported over three random seeds. 

\subsubsection{ReID Evaluation Metrics}

We use the standard re-id metrics: 
cumulative matching characteristic (CMC) rank-1, rank-5, rank-10
and mean Average Precision (mAP). 
CMC measures rank-$k$ accuracy, the probability of the true match appearing in the top $k$. 
The mAP metric computes mean average precision across all queries. 
Both operate directly on the re-id model output.

\section{Results and Discussion}

\subsection{Text Spotting}

\begin{table}[t]
\small
\centering
\caption{The RnD text detection and recognition performance 
on the RnD test set using off-the-shelf (OTS) and fine-tuned (FT) models. 
}
\label{tab:ts-results}
\begin{tabular}{@{}cccccccc@{}}
\toprule
\multicolumn{2}{c}{Model}                         & Det-P & Det-R & Det-F1 & E2E-P & E2E-R & E2E-F1 \\ \midrule
\multirow{2}{*}{OTS}        & SwinTS &   0.195    &  0.287     &  0.232      &  0.101     & 0.148      & 0.120       \\
                                      & YAMTS   &  0.192     &  0.491     &  0.276      &   0.106    &  0.244     &  0.148      \\ \midrule
\multirow{2}{*}{FT}           & SwinTS &  0.810     &  0.673     &  0.734      &   0.513    & \textbf{0.415}      &   0.459     \\
                                      & YAMTS   &  \textbf{0.847}     &  \textbf{0.715}     &  \textbf{0.775}      &    \textbf{0.758}   & 0.404      &  \textbf{0.527}      \\ \bottomrule
\end{tabular}
\end{table}

Table~\ref{tab:ts-results} summarizes the quantitative results 
on the RnD test set. 
The off-the-shelf SwinTS and YAMTS models, 
which were 
pretrained on large generic OCR datasets, 
achieve poor accuracy. 
This is expected due to
the substantial domain gap 
between existing datasets and this new motorsports application. 
Even state-of-the-art models fail 
without adaptation to racer numbers.
Fine-tuning the pretrained models on RnD 
led to major improvements. 
SwinTS achieved 0.734 detection F1 
and 0.459 end-to-end recognition F1 after fine-tuning. 
For YAMTS, 
fine-tuning improved to 0.775 detection 
and 0.527 recognition F1 scores. 
However, these fine-tuned results still 
leave substantial room for improvement.

\subsubsection{Text Spotting Performance Among Occlusion}

\begin{table}[]
\small
\centering
\caption{The RnD Performance broken down by occlusion type.}
\label{tab:occlusion_results}
\begin{tabular}{@{}lccccccc@{}}
\toprule
\multicolumn{2}{c}{Occlusion (\%)}    & Det-P & Det-R & Det-F1 & E2E-P & E2E-R & E2E-F1 \\ \midrule
\multirow{2}{*}{None (41\%)}  & OTS & 0.196 & 0.568 & 0.291  & 0.124 & 0.330 & 0.180  \\
                              & FT    & 0.880 & 0.726 & 0.795  & 0.826 & 0.470 & 0.599  \\ \midrule
\multirow{2}{*}{Blur (3\%)}   & OTS & 0.231 & 0.545 & 0.324  & 0.140 & 0.295 & 0.190  \\
                              & FT    & 0.860 & 0.841 & 0.851  & 0.750 & 0.409 & 0.529  \\ \midrule
\multirow{2}{*}{Shadow (7\%)} & OTS & 0.144 & 0.536 & 0.227  & 0.033 & 0.107 & 0.050  \\
                              & FT    & 0.875 & 0.778 & 0.824  & 0.769 & 0.370 & 0.500  \\ \midrule
\multirow{2}{*}{Mud (44\%)} & OTS & 0.194 & 0.389 & 0.259  & 0.086 & 0.152 & 0.110  \\
                              & FT    & 0.811 & 0.718 & 0.761  & 0.681 & 0.359 & 0.470  \\ \midrule
\multirow{2}{*}{Glare (8\%)}  & OTS & 0.162 & 0.547 & 0.250  & 0.052 & 0.156 & 0.078  \\
                              & FT    & 0.787 & 0.686 & 0.733  & 0.519 & 0.200 & 0.289  \\ \midrule
\multirow{2}{*}{Dust (2\%)}   & OTS & 0.173 & 0.310 & 0.222  & 0.113 & 0.190 & 0.142  \\
                              & FT    & 0.925 & 0.638 & 0.755  & 0.833 & 0.259 & 0.395  \\ \bottomrule
\end{tabular}
\end{table}

\begin{figure*}[h]
    \centering
    \begin{subfigure}[b]{0.13\textwidth}
        \centering
        \includegraphics[width=\textwidth]{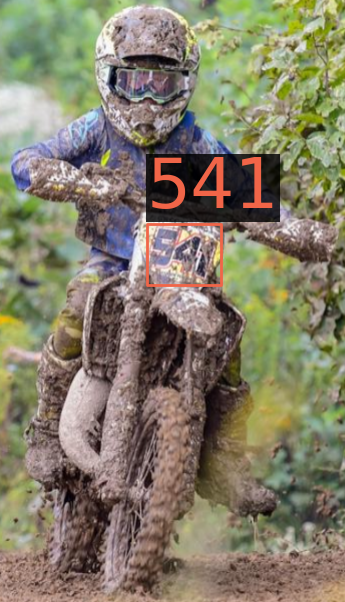}
        \caption{}
        \label{img:rtm2}
    \end{subfigure}
    \hfill
    \begin{subfigure}[b]{0.262\textwidth}
        \centering
        \includegraphics[width=\textwidth]{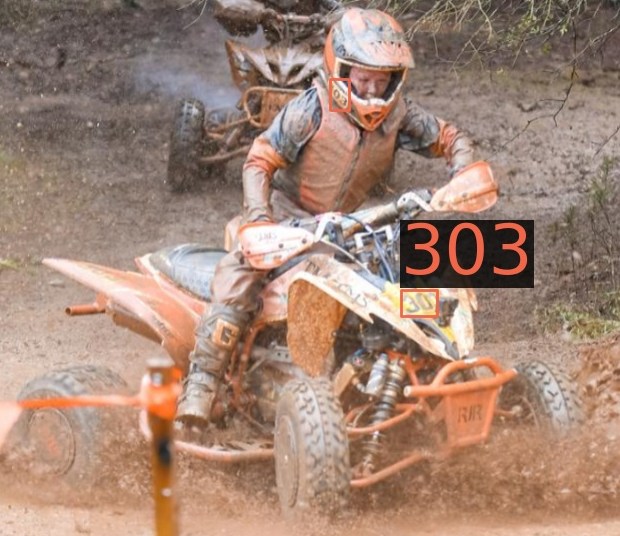}
        \caption{}
        \label{img:rtm1}
    \end{subfigure}
    \hfill
    \begin{subfigure}[b]{0.163\textwidth}
        \centering
        \includegraphics[width=\textwidth]{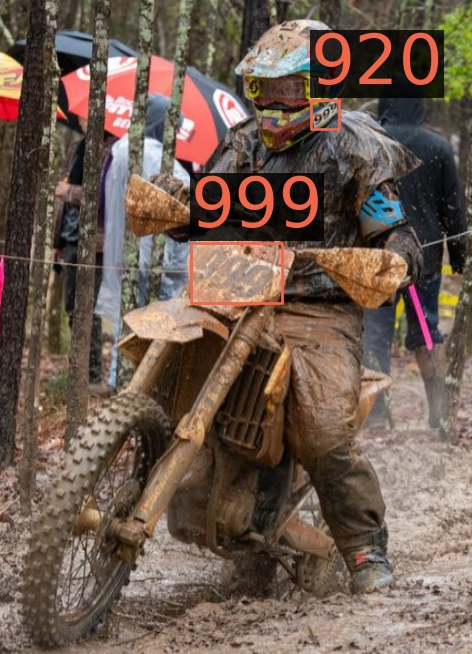}
        \caption{}
        \label{img:mudfail}
    \end{subfigure}
    \hfill
    \begin{subfigure}[b]{0.22\textwidth}
        \centering
        \includegraphics[width=\textwidth]{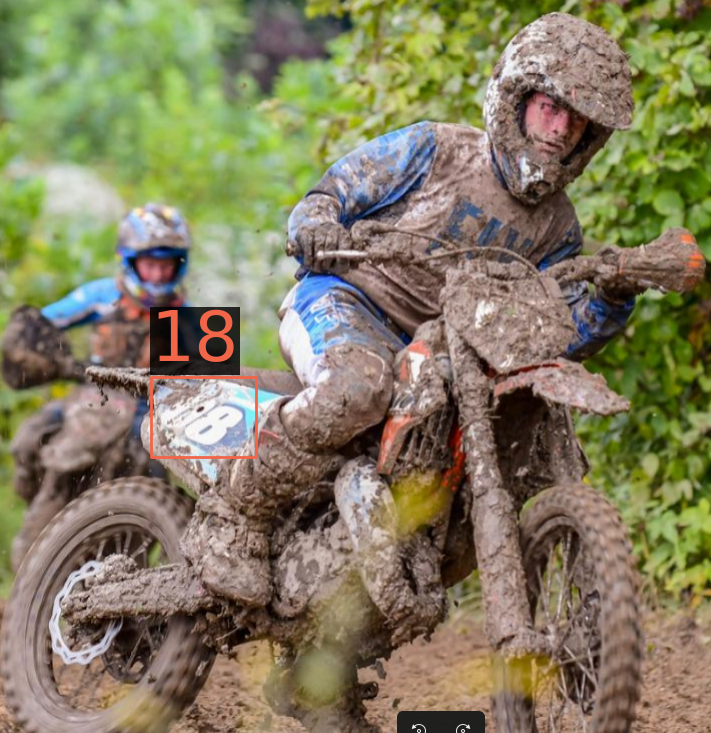}
        \caption{}
        \label{img:mudslidefail}
    \end{subfigure}
    \hfill
    \begin{subfigure}[b]{0.202\textwidth}
        \centering
        \includegraphics[width=\textwidth]{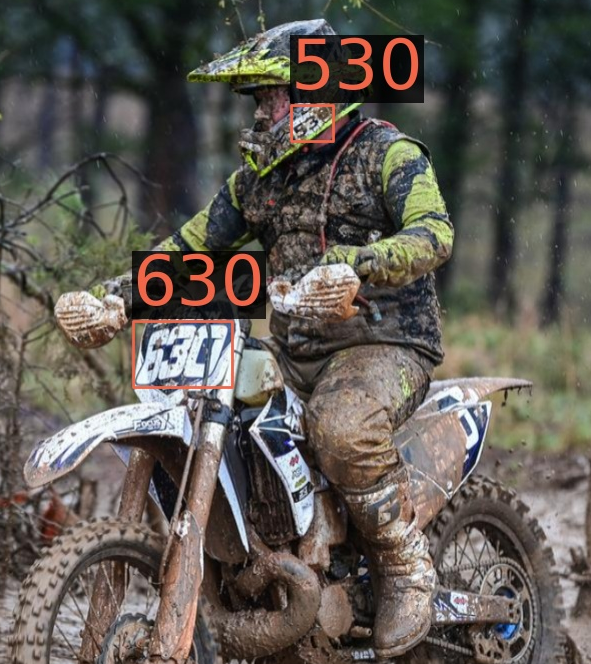}
        \caption{}
        \label{img:muddydude}
    \end{subfigure}
    \caption{Analysis of model performance on mud occluded numbers. (a) The model correctly recognizes the front number by ignoring mud. (b) The quad number is recognized but the muddy helmet number is missed. (c) The front number is read but a very muddy helmet number is missed. (d) The number is detected but misrecognized due to its odd position. (e) Two numbers are correctly read but the muddy side number is missed.}  
    \label{fig:mudfails}
\end{figure*}

We further analyzed model performance on the RnD test set 
when numbers were occluded by different factors. 
Note that a single image can contain multiple occlusions 
(i.e. it can be dusty and have glare, or it can be blurry and muddy, etc.). 
Table \ref{tab:occlusion_results} breaks down the detection 
and recognition results on images with 
no occlusion, motion blur, shadows, mud, glare, and dust.

Mud occlusion was the most prevalent, accounting for 44\% 
of the test data. 
Both off-the-shelf and fine-tuned models 
struggled with heavy mud. 
The fine-tuned model improved over the off-the-shelf version, 
achieving 0.761 detection F1 and 0.470 recognition F1 
on muddy images. 
But this remains far below the 0.795 detection 
and 0.599 recognition scores attained 
on non-occluded data. 
There is substantial room to improve 
robustness to real-world mud and dirt occlusion.

The fine-tuned model also struggled with glare occlusion, 
scoring just 0.733 detection F1 
and 0.289 recognition F1 on such images. 
Glare creates low-contrast regions that 
likely hurt feature extraction. 
Shadows likewise proved challenging, 
with a 0.824 detection but 
only a 0.500 recognition F1 score after fine-tuning. 
The changing lighting and hues may degrade recognition.

For motion blur, the fine-tuned model achieved 0.851 detection F1 
but 0.529 recognition F1. 
Blurring degrades the crispness of text features needed 
for accurate recognition. 
Surprisingly, the model performed worst on dust occlusion, 
despite it being visually less severe than mud and glare. 
This highlights the brittleness of vision models to unusual textures.

Overall, 
the breakdown reveals mud as the primary challenge, 
but substantial room remains to improve OCR accuracy 
under real-world conditions 
like shadows, dust, blur, and glare. 
Researchers should prioritize occlusions seen 
in natural operating environments 
that undermine off-the-shelf models.

\subsubsection{Text Spotting Qualitative Analysis}

\begin{figure}[h]
    \centering
    \begin{subfigure}[b]{0.47\textwidth}
        \centering
        \includegraphics[width=\textwidth]{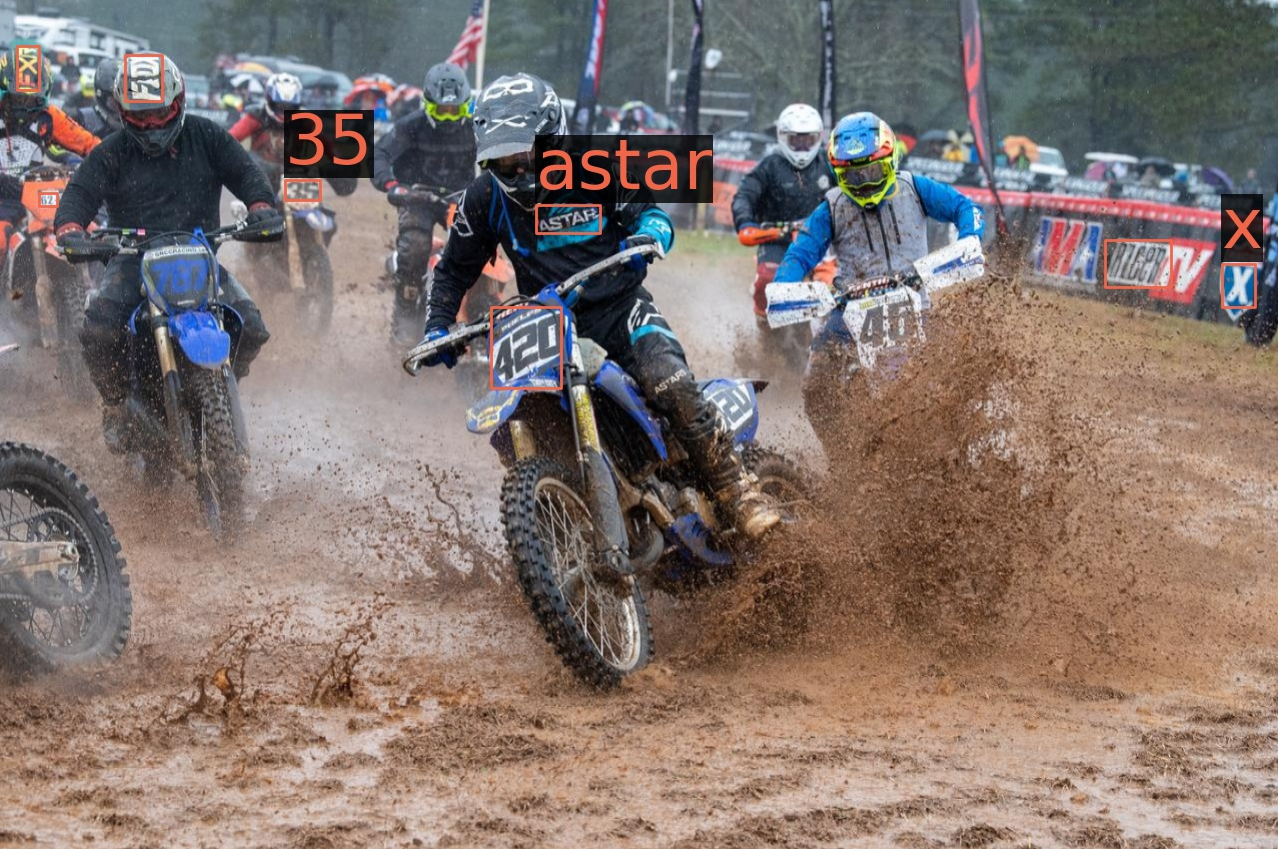}
    \end{subfigure}
    \hfill  
    \begin{subfigure}[b]{0.47\textwidth}
        \centering
        \includegraphics[width=\textwidth]{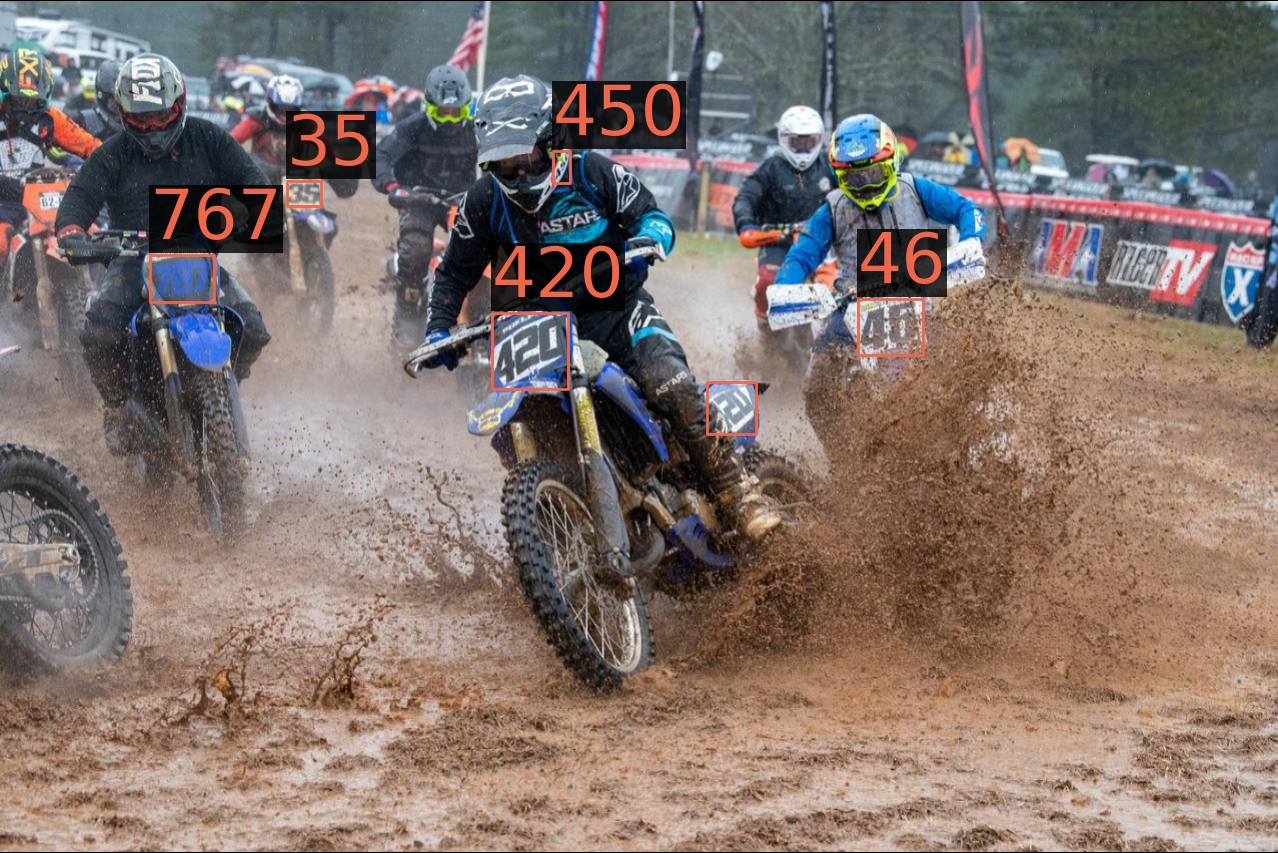}
    \end{subfigure}
    \caption{Example showcasing off-the-shelf (top image) vs fine-tuned (bottom image) model predictions in rainy conditions.}
    \label{fig:rainy-start}
\end{figure}

In this subsection, we provide a qualitative analysis of model performance on RnD using 
the fine-tuned YAMTS model, 
which achieved the highest end-to-end F1 score. 
The detection confidence threshold was set to 0.65 
and the recognition threshold is set to 0.45. 
Figures~\ref{fig:mudfails} and \ref{fig:rainy-start} 
showcase successes and failures on challenging examples.
Photos from the beginning of a race are typically the most complex,
due to the number of motorcycles in a single image and background clutter. 
Figure~\ref{fig:rainy-start} looks at a photo from the 
start of a race, but this time in rainy conditions. 
The top photo highlights the detection's of the off-the-shelf model before fine-tuning, where it can recognize only a single number properly. 
However, after fine-tuning, the model can properly recognize 5 of the 6 visible numbers.

Figure~\ref{fig:rainy-start} compares the text spotting performance before and after fine-tuning on a photo of the start of a muddy race. 
The fine-tuned model properly detects all 8 visible numbers, demonstrating capabilities to handle partial mud occlusion. 
However, it only correctly recognizes 3 of the 8 numbers, highlighting limitations in recognizing degraded text. 
Without fine-tuning, only 1 number is detected, and no numbers are properly recognized, showing the benefits of fine-tuning.
But substantial challenges remain in muddy conditions.

Figure~\ref{fig:mudfails} showcases common mud-related successes and failures. 
In some cases, the fine-tuned models can see through mud occlusions to properly recognize the racer number, as shown in Figure~\ref{img:rtm2}.
However, mud often prevents smaller helmet numbers from being recognized (Fig~\ref{img:rtm1}, \ref{img:mudfail}). 
Odd orientations also confuse models (Fig~\ref{img:mudslidefail}). 
Overall, heavy mud occlusion remains the biggest challenge.
Some other failure types, such as overlapping numbers, are discussed in Appendix~\ref{sec:additional-ts-results}.
These results show promising capabilities but also expose key areas for improvement.

\subsection{Re-ID Results}
\label{sec:results}

\begin{table*}[t]
\centering
\caption{MUDD re-id benchmark results comparing off-the-shelf, from scratch, and fine-tuning training strategies. Fine-tuning provides major accuracy gains indicating the importance of transfer learning.}
\label{tab:reid-results}
\begin{tabular}{@{}cccccc@{}}
\toprule
Training                    & Backbone  & R1 & R5 & R10 & mAP \\ \midrule
\multirow{2}{*}{Best off-the-shelf}            & OSNet     & 0.3252 & 0.5219 & 0.6327 & 0.3853 \\
                                        & ResNet-50 & 0.3164 & 0.5099 & 0.6306 & 0.3634 \\ \midrule 
\multirow{2}{*}{Trained From Scratch}        & OSNet &  0.2146 (0.03108)  &  0.4846 (0.04466)  &  0.6755 (0.03606)   &   0.2491 (0.0163)  \\
                            & ResNet-50    &  0.1591 (0.01154)  &  0.4155 (0.01564)  &  0.6194 (0.2826)   &  0.1923 (0.01853)   \\ \midrule
\multirow{2}{*}{Pretrained on Imagenet}      & OSNet &  0.7844 (0.01284)  &  0.9416 (0.005594)  &   0.9771 (0.004829)  &  0.8215 (0.01258)   \\
                            & ResNet-50    &  0.762 (0.00817)  &  0.9442 (0.004079)  &  0.9787 (0.002729)   &  0.8073 (0.006272)   \\
\multirow{2}{*}{Pretrained on MSMT17}     & OSNet &  0.7924 (0.009929)  &  0.9445 (0.001521)  &  0.9779 (0.002051)   & \textbf{0.8287 (0.005843)}    \\
                            & ResNet-50    &  0.7596 (0.02279)  & 0.9407 (0.0118)   &  0.9767 (0.006813)   &   0.8028 (0.02378)  \\ 
\multirow{2}{*}{Pretrained on DukeMTMC}     & OSNet &  0.7887 (0.01515)  &  0.9388 (0.003319)  &  97.57 (0.004367)   & 0.826 (0.0117)    \\
                            & ResNet-50    &  0.7858 (0.01726)  &  \textbf{0.9562 (0.007937)}  &  \textbf{0.9847 (0.002225)}   &   0.8277 (0.01079)  \\
\multirow{2}{*}{Pretrained on Market-1501}     & OSNet &  \textbf{0.7931 (0.01738)}  &  0.9442 (0.006051)  &  0.9778 (0.00483)   &  0.827 (0.01546)   \\
                            & ResNet-50    &  0.7812 (0.02546)  &  0.9475 (0.01395)  &  0.9807 (0.008821)   &   0.8233 (0.02077)  \\ \bottomrule
\end{tabular}
\end{table*}

Table \ref{tab:reid-results} summarizes re-id performance on MUDD.
Applying pre-trained re-id models directly to MUDD leads to very poor accuracy. 
The highest Rank-1 is only 32.52\% using OSNet pre-trained on Market-1501. 
Again, this is expected as there is a significant domain gap between existing re-id datasets and MUDD's challenging conditions. 
Yet training from scratch proves even more difficult, yielding only 21.46\% rank-1 accuracy. 
Fine-tuning pre-trained models on MUDD proves the most effective strategy. 
Fine-tuned OSNet reaches 79.31\% Rank-1, over 2.5x higher than the best off-the-shelf model. 

Interestingly, models pre-trained on generic ImageNet data perform nearly as well as those pre-trained on re-id specific datasets, like Market-1501, after fine-tuning. 
This is likely because MUDD represents a significant domain shift even from existing re-id datasets, yet we are still able to glean benefits from generic pretraining. 
In terms of architecture, we find that the OSNet-based models achieve slightly higher rank-1 accuracy (79.31\%) versus the more traditional ResNet-50 (78.12\%). While OSNet is a specialized architecture for person re-identification, ResNet's more general application still performs competitively. 
Overall, both architectures can adapt to MUDD's domain when fine-tuned, with OSNet's inductive biases providing a small boost. 

We also were interested in determining the right pretraining corpora to get the best-performing MUDD model. 
We compared the models pretrained on one of the 
re-id datasets of MSMT17~\citep{wei2018person}, DukeMTMC~\citep{ristani2016performance}, or Market-1501~\citep{zheng2015scalable}, which are all then further fine-tuned on MUDD. 
The performance of these models is nearly the same across different source datasets.
They all substantially improve over off-the-shelf and from scratch approaches, but don't present any particular benefit over one another.

\subsubsection{Analysis}

\begin{table}[]
\caption{The performance of the best ReID method when controlling the query and gallery set for muddy images. ``No Mud -\textgreater Mud'' corresponds to the query set containing only clean images, and the gallery set containing only muddy images.}
\label{tab:reid-mudanalysis}
\begin{tabular}{@{}ccccc@{}}
\toprule
Query Set -\textgreater \ Gallery Set  & Rank1  & Rank5  & Rank10 & mAP    \\ \midrule
No Mud -\textgreater \ No Mud          & 0.8852 & 0.9727 & 0.9968 & 0.8809 \\
Mud -\textgreater \ Mud                & 0.5624 & 0.7162 & 0.7349 & 0.6002 \\
No Mud -\textgreater \ Mud             & 0.7342 & 0.8641 & 0.8543 & 0.6916 \\
Mud -\textgreater \ No Mud             & 0.7335 & 0.8267 & 0.8333 & 0.7690 \\ \bottomrule
\end{tabular}
\end{table}

\begin{figure*}[t]
    \centering
    \includegraphics[width=\textwidth]{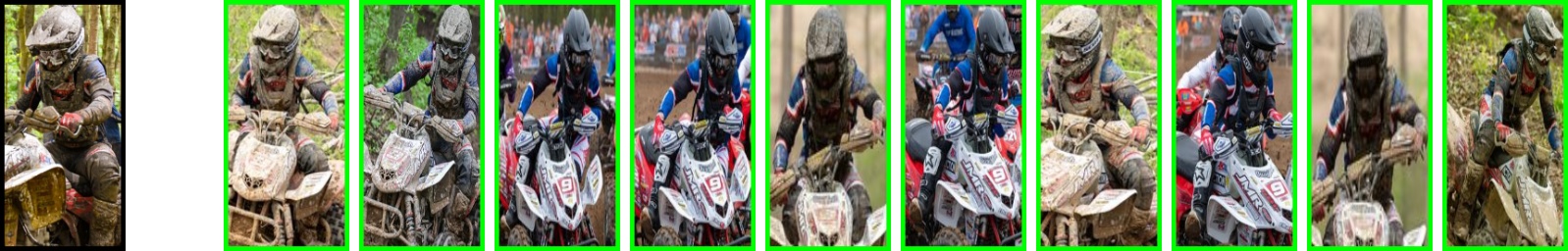}
    \caption{Example of successful re-id by the fine-tuned model under light mud occlusion. All top 10 ranked results correctly match the query rider despite the mud, blurring, lighting, pose, and complex backgrounds. Green boundaries signify correct matches and red incorrect.}
    \label{fig:light-mud-success}
\end{figure*}

\begin{figure*}[t]
    \centering
    \includegraphics[width=\textwidth]{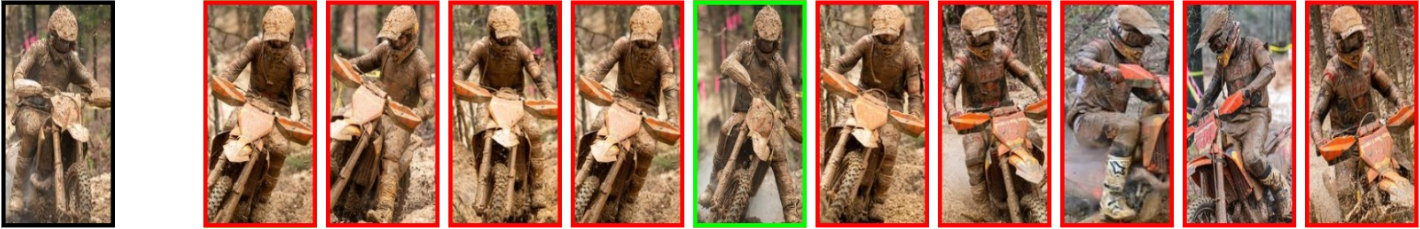}
    \caption{Failure case with heavy mud occlusion on the query image. Only 1 out of the top 10 results is a correct match, despite over 20 images of the same rider appearing in the gallery set, most of which are clean. Green boundaries signify correct matches and red incorrect.
    }
    \label{fig:very-muddy-query}
\end{figure*}

Our fine-tuned models demonstrate significant improvements in re-identifying riders compared to off-the-shelf and random initialization from scratch approaches. 
As seen in Figure~\ref{fig:light-mud-success}, the model can correctly match identities even with significant mud occlusion. 
This indicates that fine-tuning successfully incorporates invariances to mud while still distinguishing small inter-class differences like the shape of a rider's gear.
However, challenges remain under more extreme conditions. 
In the rest of this section, we analyze several key factors that still cause fine-tuned model failures on MUDD.

As expected, heavy mud occlusion poses the most significant challenges. 
And while we do see significant progress and capabilities on this front, there is more room for improvement.
Mud induces high intra-class variation as the amount of mud covering a rider can vary drastically across images. 
It also causes low inter-class variation since mud occludes distinguishing features like jersey numbers and colors. 
As shown in Figure \ref{fig:very-muddy-query}, querying with a muddy image typically retrieves other muddy images rather than cleaner images of the same identity.

Appendix~\ref{sec:additional-reid-results} includes more failure cases and examples. 
In summary, we see that changes in the natural appearance of a rider over a race also confuse models. 
Riders may change gear like goggles or gloves multiple times. 
Crashes can rip clothing and jerseys. 
The model must learn to link different levels of mud, gear, and damage to a rider.
Complex poses like jumps, crashes, and wheelies are difficult to match, especially combined with mud and appearance variation. 

Table~\ref{tab:reid-mudanalysis} breaks down the performance of the 
best fine-tuned re-id model by controlling the query and gallery set for muddy and clean images.
We see that by looking only at the clean imagery (i.e. no mud), we get a much better performing model, with gains around 10\%. 
On the other hand, when we evaluate using only the muddy imagery, we see drops in performance of around 20\% across the board. 
Lastly, when a clean or muddy image is used as the query point, and the opposite is used for the gallery set, performance falls between the evaluation using only muddy or only clean images. 
The accuracy is a bit higher for matching muddy query images to clean gallery images versus the reverse.

\section{Related Work}

\subsection{Person Re-Identification}

Person re-identification (re-id) 
aims to match people across 
non-overlapping camera views and time 
horizons~\citep{ye2021deep, zheng2016person, zheng2015scalable, farenzena2010person}. 
Early re-id methods relied on handcrafted features 
like color histograms, textures, 
and local descriptors~\citep{farenzena2010person}. 
With the rise of deep learning, 
Convolutional Neural Network (CNN)~\cite{zhou2019omni} 
and Transformer~\citep{he2021transreid} 
based approaches now dominate re-id research, 
spurred by datasets like Market-1501~\citep{zheng2015scalable}, 
DukeMTMC-ReID~\citep{ristani2016performance}, 
and MSMT17~\citep{wei2018person}.

A few datasets address environmental factors. 
For example, 
\citet{xiao2016end} introduce a dataset with a low-resolution 
challenge set. 
Occlusions have also 
been well studied, 
spearheaded by datasets with high levels of occlusion~\citep{schwartz2009learning, wang2011re, wang2016person, figueira2015hda+, xiao2016end}.
However, 
these occlusions are unrelated
to the heavy mud occlusion in our dataset. 
The addition of mud drastically 
complicates re-identification.
Furthermore, 
no prior datasets exhibit such a complex combination of lighting, 
diversity, motion, and diverse cameras as 
our off-road racing dataset. 

Prior work has focused on the re-identification of motorcycles and bicycles~\cite{figueiredo2021more, li2022rider, yuan2018bike}, 
however, these are restricted to street vehicles in urban settings.
A highly related domain is identifying athletes in sports imagery.
\citet{PENATESANCHEZ2020355} released a dataset of ultra-runners
competing in a 128km race over the course of a day and a night. 
While this is more similar to the off-road setting in our dataset, 
they only have 416 different identities between 5 locations at a single event. 
Furthermore, 
there is near zero mud in the dataset.
Along similar lines, 
but in even more controlled and limited settings, 
are the SoccarNet-ReID~\citep{giancola2022soccernet} 
and DeepSportRadar-ReID~\citep{van2022deepsportradar} datasets, 
which contain images from broadcast video 
of soccer and basketball games respectively. 

These datasets have driven research to develop 
methods to deal with the occlusions common in them. 
Approaches such as 
invariant representations~\citep{chen2019learning}, 
metric learning~\citep{yi2014deep}, 
semantic attributes~\citep{shi2015transferring},
part-based~\citep{cheng2016person} 
and pose-aware models~\cite{cho2016improving}, 
and adversarial learning~\citep{huang2018adversarially}
have been proposed to alleviate occlusion problems. 
Other methods have been developed to 
handle misalignment,
utilizes temporal cues in video~\citep{li2019global},
use domain adaptation techniques~\citep{deng2018image},
or unsupervised methods~\citep{fan2018unsupervised}
to reduce label dependencies. 
Unlike our dataset, 
these all operate in controlled conditions.
Existing models thus fail on our data. 

In summary, 
re-id research has focused 
on controlled conditions 
and modest variation. 
Our dataset introduces real-world challenges 
absent in existing datasets.
Our experiments expose clear gaps 
between current methods and this
application. 
MUDD provides diverse imagery to spur 
new techniques for robust re-id under uncontrolled conditions.

\subsection{Text Spotting}

Text detection and recognition in images 
is a classic computer vision task. 
Early traditional methods relied on sliding windows, 
connected components, 
and handcrafted features like HOG~\cite{wang2010word}. 
With the advent of deep learning, 
convolutional and recurrent neural networks 
now dominate scene text recognition pipelines~\cite{textinthewild}. 
Models leverage large annotated datasets 
to learn powerful representations tuned for 
text detection and recognition in a specific domain.

Many datasets and competitions have driven progress 
in general OCR. 
These include ICDAR~\cite{karatzas2013icdar}, 
COCO-Text~\cite{lin2014microsoft}, 
and Street View Text~\cite{wang2011end}.
Popular detection models build on 
Region Proposal Networks 
and include CTPN~\cite{tian2016detecting}, 
EAST~\cite{Zhou_2017_CVPR}, 
and Craft~\cite{baek2019character}. 
Recognition is often achieved via CNN + RNN architectures 
like CRNN~\cite{CRNN} 
or transformer networks like ASTER~\cite{aster}. 
More recent state-of-the-art methods utilize 
pre-trained vision models like 
ViTSTR~\cite{vitstr}, 
PARSeq~\cite{bautista2022parseq}, 
CLIP4STR~\cite{zhao2023clip4str},
and DeepSolo~\cite{Ye_2023_CVPR}. 
However, 
most OCR research targets 
images of documents, 
signs, or web images. 
While many of these works aim to go beyond structured settings 
(e.g.images of documents, signs, or web images)
and address the task of ``robust reading'', i.e. OCR in incidental or real scenes,
recognizing text ``in the wild''
with few assumptions remains an open challenge~\cite{textinthewild}.
Furthermore, 
domain gaps exist where current methods fail 
on specialized applications. 
Our work focuses on one such gap - 
recognizing racer numbers in motorsports.

A few prior works address detecting 
and recognizing 
the license plates on vehicles~\cite{ap2020automatic, laroca2018robust, chen2019automatic, lee2019snider, silva2018license, quang2022character, laroca2021efficient}.
Some have focused specifically on street motorcycle 
number plates~\cite{kulkarni2018automatic, sathe2022helmet, sanjana2021review, lee2004extraction}.
All of these efforts use data gathered from some form of 
street camera, 
which are placed in strategic locations 
with recognizing license plates specifically in mind. 
In contrast, 
our dataset is gathered from 
professional motorsport photographers 
focused on capturing the most aesthetically pleasing 
photograph of each racer. 
Furthermore, 
existing datasets have standardized plates 
which differ greatly 
from the diverse layouts and occlusions 
of off-road motorcycle numbers. 
Street motorcycle plates exhibit 
consistency in position and appearance, 
unlike the numbers encountered during off-road competitions. 
The conditions during races also introduce 
and exacerbate factors like motion blur, 
mud occlusion, 
glare, 
and shaky cameras are not prevalent in street imagery. 
RnD provides novel real-world imagery to push OCR capabilities.

The most relevant prior domain 
is recognizing runner bib numbers 
in marathon images~\cite{shivakumara2017new, ben2012racing, boonsim2018racing, kamlesh2017person}. 
This shares similarities, 
but runner bibs provide 
more spatial and appearance consistency 
than motorcycle racing numbers. 
Datasets like TGCRBNW~\cite{TGCRBNW} 
exhibit some motion blur and night racing, 
but do not contain the mud, 
vehicle occlusion and diversity of layouts seen in motorsports.

Number recognition has also been studied in other sports - 
football~\cite{yamamoto2013multiple, bhargavi2022knock}, 
soccer~\cite{Gerke_2015_ICCV_Workshops, gerke2017soccer, vsaric2008player, diop2022soccer, alhejaily2023automatic}, 
basketball~\cite{ahammed2018basketball}, 
track and 
field~\cite{messelodi2013scene}, 
and more~\cite{liu2019pose, nag2019crnn, vats2021multi, wronska2017athlete}.
However, most focus on jersey numbers in 
commercial broadcast footage rather 
than track/field-side imagery. 
Existing sports datasets offer limited diversity and size. 
To our knowledge, 
RnD represents the largest and most varied collection of motorsports numbers in natural contexts.

In summary, 
prior work has made great progress in OCR 
for documents, signs, and other domains,
but real-world applications like 
recognizing racers in off-road competitions 
remain extremely challenging due to domain gaps in current data. 
RnD provides novel imagery to spur advances in OCR for motorsports. 
Our benchmark experiments expose substantial room 
for improvement using this data.

\section{Conclusion}

In this work, we introduce two datasets jointly referred to as Beyond the Mud. 
These two datasets, RnD and MUDD, are the first datasets to benchmark 
the performance of modern methods for text spotting and person re-identification in off-road motorcycle races. 
These datasets capture challenging factors including heavy mud occlusion, complex poses, variable lighting, occlusions, complex backgrounds, monition blur, and more. 
All images were collected from \url{PerformancePhoto.co}, and captured by 16 professional photographers across 50 distinct off-road competitions, distributed geographically throughout the USA. 
To our knowledge, these datasets represent the largest, most varied collection of annotated motorsports numbers and racers in unconstrained environments.

Benchmark results on both datasets highlight significant room for improvement. 
This dataset is significantly different than typical, leading to extremely low off-the-shelf (i.e. transfer learning) performance. 
The best off-the-shelf text-spotting method achieved an end-to-end F1 score of only 15\%, and the best off-the-shelf person re-identification method reached only 33\% rank-1 accuracy. 
When fine-tuning models to these datasets, the extremely high intra-class variability, paired with the low inter-class variability makes training particularly difficult. The best text-spotting model achieves an end-to-end F1 score of only 53\%, while the best person re-identification model produces a rank-1 accuracy of 79\%. 

Through both quantitative and qualitative analysis, we revealed some of the primary factors degrading performance in both settings to be heavy mud occlusion, glare, dust, and complex poses. 
Our work exposes the limitations of current models and methods amongst extreme off-road conditions and provides datasets of real-world imagery with baseline results. 
We hope the community will build upon these initial experiments to make advances on the problem of accurately reading text and matching racers in extreme environments.

\bibliographystyle{ACM-Reference-Format}
\bibliography{combined-refs}


\begin{thebibliography}{83}


\ifx \showCODEN    \undefined \def \showCODEN     #1{\unskip}     \fi
\ifx \showDOI      \undefined \def \showDOI       #1{#1}\fi
\ifx \showISBNx    \undefined \def \showISBNx     #1{\unskip}     \fi
\ifx \showISBNxiii \undefined \def \showISBNxiii  #1{\unskip}     \fi
\ifx \showISSN     \undefined \def \showISSN      #1{\unskip}     \fi
\ifx \showLCCN     \undefined \def \showLCCN      #1{\unskip}     \fi
\ifx \shownote     \undefined \def \shownote      #1{#1}          \fi
\ifx \showarticletitle \undefined \def \showarticletitle #1{#1}   \fi
\ifx \showURL      \undefined \def \showURL       {\relax}        \fi
\providecommand\bibfield[2]{#2}
\providecommand\bibinfo[2]{#2}
\providecommand\natexlab[1]{#1}
\providecommand\showeprint[2][]{arXiv:#2}

\bibitem[Ahammed(2018)]%
        {ahammed2018basketball}
\bibfield{author}{\bibinfo{person}{Zubaer Ahammed}.}
  \bibinfo{year}{2018}\natexlab{}.
\newblock \emph{\bibinfo{title}{Basketball player identification by jersey and
  number recognition}}.
\newblock \bibinfo{thesistype}{Ph.\,D. Dissertation}. \bibinfo{school}{Brac
  University}.
\newblock


\bibitem[Alhejaily et~al\mbox{.}(2023)]%
        {alhejaily2023automatic}
\bibfield{author}{\bibinfo{person}{Ragd Alhejaily}, \bibinfo{person}{Rahaf
  Alhejaily}, \bibinfo{person}{Mai Almdahrsh}, \bibinfo{person}{Shareefah
  Alessa}, {and} \bibinfo{person}{Saleh Albelwi}.}
  \bibinfo{year}{2023}\natexlab{}.
\newblock \showarticletitle{Automatic Team Assignment and Jersey Number
  Recognition in Football Videos}.
\newblock \bibinfo{journal}{\emph{INTELLIGENT AUTOMATION AND SOFT COMPUTING}}
  \bibinfo{volume}{36}, \bibinfo{number}{3} (\bibinfo{year}{2023}),
  \bibinfo{pages}{2669--2684}.
\newblock


\bibitem[Ap et~al\mbox{.}(2020)]%
        {ap2020automatic}
\bibfield{author}{\bibinfo{person}{N~Palanivel Ap}, \bibinfo{person}{T
  Vigneshwaran}, \bibinfo{person}{M~Sriv Arappradhan}, {and} \bibinfo{person}{R
  Madhanraj}.} \bibinfo{year}{2020}\natexlab{}.
\newblock \showarticletitle{Automatic number plate detection in vehicles using
  faster R-CNN}. In \bibinfo{booktitle}{\emph{2020 International conference on
  system, computation, automation and networking (ICSCAN)}}. IEEE,
  \bibinfo{pages}{1--6}.
\newblock


\bibitem[Appalaraju et~al\mbox{.}(2021)]%
        {Appalaraju_2021_ICCV}
\bibfield{author}{\bibinfo{person}{Srikar Appalaraju}, \bibinfo{person}{Bhavan
  Jasani}, \bibinfo{person}{Bhargava~Urala Kota}, \bibinfo{person}{Yusheng
  Xie}, {and} \bibinfo{person}{R. Manmatha}.} \bibinfo{year}{2021}\natexlab{}.
\newblock \showarticletitle{DocFormer: End-to-End Transformer for Document
  Understanding}. In \bibinfo{booktitle}{\emph{Proceedings of the IEEE/CVF
  International Conference on Computer Vision (ICCV)}}.
  \bibinfo{pages}{993--1003}.
\newblock


\bibitem[Atienza(2021)]%
        {vitstr}
\bibfield{author}{\bibinfo{person}{Rowel Atienza}.}
  \bibinfo{year}{2021}\natexlab{}.
\newblock \showarticletitle{Vision Transformer for Fast and Efficient Scene
  Text Recognition}. In \bibinfo{booktitle}{\emph{Document Analysis and
  Recognition -- ICDAR 2021}}, \bibfield{editor}{\bibinfo{person}{Josep
  Llad{\'o}s}, \bibinfo{person}{Daniel Lopresti}, {and}
  \bibinfo{person}{Seiichi Uchida}} (Eds.). \bibinfo{publisher}{Springer
  International Publishing}, \bibinfo{address}{Cham},
  \bibinfo{pages}{319--334}.
\newblock
\showISBNx{978-3-030-86549-8}


\bibitem[Baek et~al\mbox{.}(2019)]%
        {baek2019character}
\bibfield{author}{\bibinfo{person}{Youngmin Baek}, \bibinfo{person}{Bado Lee},
  \bibinfo{person}{Dongyoon Han}, \bibinfo{person}{Sangdoo Yun}, {and}
  \bibinfo{person}{Hwalsuk Lee}.} \bibinfo{year}{2019}\natexlab{}.
\newblock \showarticletitle{Character region awareness for text detection}. In
  \bibinfo{booktitle}{\emph{Proceedings of the IEEE/CVF conference on computer
  vision and pattern recognition}}. \bibinfo{pages}{9365--9374}.
\newblock


\bibitem[Bautista and Atienza(2022)]%
        {bautista2022parseq}
\bibfield{author}{\bibinfo{person}{Darwin Bautista} {and}
  \bibinfo{person}{Rowel Atienza}.} \bibinfo{year}{2022}\natexlab{}.
\newblock \showarticletitle{Scene Text Recognition with Permuted Autoregressive
  Sequence Models}. In \bibinfo{booktitle}{\emph{European Conference on
  Computer Vision}}. \bibinfo{publisher}{Springer Nature Switzerland},
  \bibinfo{address}{Cham}, \bibinfo{pages}{178--196}.
\newblock
\urldef\tempurl%
\url{https://doi.org/10.1007/978-3-031-19815-1_11}
\showDOI{\tempurl}


\bibitem[Ben-Ami et~al\mbox{.}(2012)]%
        {ben2012racing}
\bibfield{author}{\bibinfo{person}{Idan Ben-Ami}, \bibinfo{person}{Tali Basha},
  {and} \bibinfo{person}{Shai Avidan}.} \bibinfo{year}{2012}\natexlab{}.
\newblock \showarticletitle{Racing Bib Numbers Recognition.}. In
  \bibinfo{booktitle}{\emph{BMVC}}. \bibinfo{pages}{1--10}.
\newblock


\bibitem[Bhargavi et~al\mbox{.}(2022)]%
        {bhargavi2022knock}
\bibfield{author}{\bibinfo{person}{Divya Bhargavi},
  \bibinfo{person}{Erika~Pelaez Coyotl}, {and} \bibinfo{person}{Sia Gholami}.}
  \bibinfo{year}{2022}\natexlab{}.
\newblock \showarticletitle{Knock, knock. Who's there?--Identifying football
  player jersey numbers with synthetic data}.
\newblock \bibinfo{journal}{\emph{arXiv preprint arXiv:2203.00734}}
  (\bibinfo{year}{2022}).
\newblock


\bibitem[Boonsim(2018)]%
        {boonsim2018racing}
\bibfield{author}{\bibinfo{person}{Noppakun Boonsim}.}
  \bibinfo{year}{2018}\natexlab{}.
\newblock \showarticletitle{Racing bib number localization on complex
  backgrounds}.
\newblock \bibinfo{journal}{\emph{WSEAS Transactions on Systems and Control}}
  \bibinfo{volume}{13} (\bibinfo{year}{2018}), \bibinfo{pages}{226--231}.
\newblock


\bibitem[Chen et~al\mbox{.}(2019a)]%
        {chen2019automatic}
\bibfield{author}{\bibinfo{person}{Rung-Ching Chen} {et~al\mbox{.}}}
  \bibinfo{year}{2019}\natexlab{a}.
\newblock \showarticletitle{Automatic License Plate Recognition via
  sliding-window darknet-YOLO deep learning}.
\newblock \bibinfo{journal}{\emph{Image and Vision Computing}}
  \bibinfo{volume}{87} (\bibinfo{year}{2019}), \bibinfo{pages}{47--56}.
\newblock


\bibitem[Chen et~al\mbox{.}(2021)]%
        {textinthewild}
\bibfield{author}{\bibinfo{person}{Xiaoxue Chen}, \bibinfo{person}{Lianwen
  Jin}, \bibinfo{person}{Yuanzhi Zhu}, \bibinfo{person}{Canjie Luo}, {and}
  \bibinfo{person}{Tianwei Wang}.} \bibinfo{year}{2021}\natexlab{}.
\newblock \showarticletitle{Text Recognition in the Wild: A Survey}.
\newblock \bibinfo{journal}{\emph{ACM Comput. Surv.}} \bibinfo{volume}{54},
  \bibinfo{number}{2}, Article \bibinfo{articleno}{42} (\bibinfo{date}{mar}
  \bibinfo{year}{2021}), \bibinfo{numpages}{35}~pages.
\newblock
\showISSN{0360-0300}
\urldef\tempurl%
\url{https://doi.org/10.1145/3440756}
\showDOI{\tempurl}


\bibitem[Chen et~al\mbox{.}(2019b)]%
        {chen2019learning}
\bibfield{author}{\bibinfo{person}{Yun-Chun Chen}, \bibinfo{person}{Yu-Jhe Li},
  \bibinfo{person}{Xiaofei Du}, {and} \bibinfo{person}{Yu-Chiang~Frank Wang}.}
  \bibinfo{year}{2019}\natexlab{b}.
\newblock \showarticletitle{Learning resolution-invariant deep representations
  for person re-identification}. In \bibinfo{booktitle}{\emph{Proceedings of
  the AAAI conference on artificial intelligence}}, Vol.~\bibinfo{volume}{33}.
  \bibinfo{pages}{8215--8222}.
\newblock


\bibitem[Cheng et~al\mbox{.}(2016)]%
        {cheng2016person}
\bibfield{author}{\bibinfo{person}{De Cheng}, \bibinfo{person}{Yihong Gong},
  \bibinfo{person}{Sanping Zhou}, \bibinfo{person}{Jinjun Wang}, {and}
  \bibinfo{person}{Nanning Zheng}.} \bibinfo{year}{2016}\natexlab{}.
\newblock \showarticletitle{Person re-identification by multi-channel
  parts-based cnn with improved triplet loss function}. In
  \bibinfo{booktitle}{\emph{Proceedings of the iEEE conference on computer
  vision and pattern recognition}}. \bibinfo{pages}{1335--1344}.
\newblock


\bibitem[Ch'ng and Chan(2017)]%
        {ch2017total}
\bibfield{author}{\bibinfo{person}{Chee~Kheng Ch'ng} {and}
  \bibinfo{person}{Chee~Seng Chan}.} \bibinfo{year}{2017}\natexlab{}.
\newblock \showarticletitle{Total-text: A comprehensive dataset for scene text
  detection and recognition}. In \bibinfo{booktitle}{\emph{2017 14th IAPR
  international conference on document analysis and recognition (ICDAR)}},
  Vol.~\bibinfo{volume}{1}. IEEE, \bibinfo{pages}{935--942}.
\newblock


\bibitem[Cho and Yoon(2016)]%
        {cho2016improving}
\bibfield{author}{\bibinfo{person}{Yeong-Jun Cho} {and}
  \bibinfo{person}{Kuk-Jin Yoon}.} \bibinfo{year}{2016}\natexlab{}.
\newblock \showarticletitle{Improving person re-identification via pose-aware
  multi-shot matching}. In \bibinfo{booktitle}{\emph{Proceedings of the IEEE
  conference on computer vision and pattern recognition}}.
  \bibinfo{pages}{1354--1362}.
\newblock


\bibitem[Deng et~al\mbox{.}(2018)]%
        {deng2018image}
\bibfield{author}{\bibinfo{person}{Weijian Deng}, \bibinfo{person}{Liang
  Zheng}, \bibinfo{person}{Qixiang Ye}, \bibinfo{person}{Guoliang Kang},
  \bibinfo{person}{Yi Yang}, {and} \bibinfo{person}{Jianbin Jiao}.}
  \bibinfo{year}{2018}\natexlab{}.
\newblock \showarticletitle{Image-image domain adaptation with preserved
  self-similarity and domain-dissimilarity for person re-identification}. In
  \bibinfo{booktitle}{\emph{Proceedings of the IEEE conference on computer
  vision and pattern recognition}}. \bibinfo{pages}{994--1003}.
\newblock


\bibitem[Diop et~al\mbox{.}(2022)]%
        {diop2022soccer}
\bibfield{author}{\bibinfo{person}{Charles-Alexandre Diop},
  \bibinfo{person}{Baptiste Pelloux}, \bibinfo{person}{Xinrui Yu},
  \bibinfo{person}{Won-Jae Yi}, {and} \bibinfo{person}{Jafar Saniie}.}
  \bibinfo{year}{2022}\natexlab{}.
\newblock \showarticletitle{Soccer Player Recognition using Artificial
  Intelligence and Computer Vision}. In \bibinfo{booktitle}{\emph{2022 IEEE
  International Conference on Electro Information Technology (eIT)}}. IEEE,
  \bibinfo{pages}{477--481}.
\newblock


\bibitem[Fan et~al\mbox{.}(2018)]%
        {fan2018unsupervised}
\bibfield{author}{\bibinfo{person}{Hehe Fan}, \bibinfo{person}{Liang Zheng},
  \bibinfo{person}{Chenggang Yan}, {and} \bibinfo{person}{Yi Yang}.}
  \bibinfo{year}{2018}\natexlab{}.
\newblock \showarticletitle{Unsupervised person re-identification: Clustering
  and fine-tuning}.
\newblock \bibinfo{journal}{\emph{ACM Transactions on Multimedia Computing,
  Communications, and Applications (TOMM)}} \bibinfo{volume}{14},
  \bibinfo{number}{4} (\bibinfo{year}{2018}), \bibinfo{pages}{1--18}.
\newblock


\bibitem[Farenzena et~al\mbox{.}(2010)]%
        {farenzena2010person}
\bibfield{author}{\bibinfo{person}{Michela Farenzena}, \bibinfo{person}{Loris
  Bazzani}, \bibinfo{person}{Alessandro Perina}, \bibinfo{person}{Vittorio
  Murino}, {and} \bibinfo{person}{Marco Cristani}.}
  \bibinfo{year}{2010}\natexlab{}.
\newblock \showarticletitle{Person re-identification by symmetry-driven
  accumulation of local features}. In \bibinfo{booktitle}{\emph{2010 IEEE
  computer society conference on computer vision and pattern recognition}}.
  IEEE, \bibinfo{pages}{2360--2367}.
\newblock


\bibitem[Figueira et~al\mbox{.}(2015)]%
        {figueira2015hda+}
\bibfield{author}{\bibinfo{person}{Dario Figueira}, \bibinfo{person}{Matteo
  Taiana}, \bibinfo{person}{Athira Nambiar}, \bibinfo{person}{Jacinto
  Nascimento}, {and} \bibinfo{person}{Alexandre Bernardino}.}
  \bibinfo{year}{2015}\natexlab{}.
\newblock \showarticletitle{The HDA+ data set for research on fully automated
  re-identification systems}. In \bibinfo{booktitle}{\emph{Computer Vision-ECCV
  2014 Workshops: Zurich, Switzerland, September 6-7 and 12, 2014, Proceedings,
  Part III 13}}. Springer, \bibinfo{pages}{241--255}.
\newblock


\bibitem[Figueiredo et~al\mbox{.}(2021)]%
        {figueiredo2021more}
\bibfield{author}{\bibinfo{person}{Augusto Figueiredo},
  \bibinfo{person}{Johnata Brayan}, \bibinfo{person}{Renan~Oliveira Reis},
  \bibinfo{person}{Raphael Prates}, {and} \bibinfo{person}{William~Robson
  Schwartz}.} \bibinfo{year}{2021}\natexlab{}.
\newblock \showarticletitle{More: a large-scale motorcycle re-identification
  dataset}. In \bibinfo{booktitle}{\emph{Proceedings of the IEEE/CVF Winter
  Conference on Applications of Computer Vision}}. \bibinfo{pages}{4034--4043}.
\newblock


\bibitem[Fujitake(2024)]%
        {fujitake2024dtrocr}
\bibfield{author}{\bibinfo{person}{Masato Fujitake}.}
  \bibinfo{year}{2024}\natexlab{}.
\newblock \showarticletitle{Dtrocr: Decoder-only transformer for optical
  character recognition}. In \bibinfo{booktitle}{\emph{Proceedings of the
  IEEE/CVF Winter Conference on Applications of Computer Vision}}.
  \bibinfo{pages}{8025--8035}.
\newblock


\bibitem[Ge et~al\mbox{.}(2021)]%
        {ge2021yolox}
\bibfield{author}{\bibinfo{person}{Zheng Ge}, \bibinfo{person}{Songtao Liu},
  \bibinfo{person}{Feng Wang}, \bibinfo{person}{Zeming Li}, {and}
  \bibinfo{person}{Jian Sun}.} \bibinfo{year}{2021}\natexlab{}.
\newblock \showarticletitle{Yolox: Exceeding yolo series in 2021}.
\newblock \bibinfo{journal}{\emph{arXiv preprint arXiv:2107.08430}}
  (\bibinfo{year}{2021}).
\newblock


\bibitem[Gerke et~al\mbox{.}(2017)]%
        {gerke2017soccer}
\bibfield{author}{\bibinfo{person}{Sebastian Gerke}, \bibinfo{person}{Antje
  Linnemann}, {and} \bibinfo{person}{Karsten M{\"u}ller}.}
  \bibinfo{year}{2017}\natexlab{}.
\newblock \showarticletitle{Soccer player recognition using spatial
  constellation features and jersey number recognition}.
\newblock \bibinfo{journal}{\emph{Computer Vision and Image Understanding}}
  \bibinfo{volume}{159} (\bibinfo{year}{2017}), \bibinfo{pages}{105--115}.
\newblock


\bibitem[Gerke et~al\mbox{.}(2015)]%
        {Gerke_2015_ICCV_Workshops}
\bibfield{author}{\bibinfo{person}{Sebastian Gerke}, \bibinfo{person}{Karsten
  Muller}, {and} \bibinfo{person}{Ralf Schafer}.}
  \bibinfo{year}{2015}\natexlab{}.
\newblock \showarticletitle{Soccer Jersey Number Recognition Using
  Convolutional Neural Networks}. In \bibinfo{booktitle}{\emph{Proceedings of
  the IEEE International Conference on Computer Vision (ICCV) Workshops}}.
\newblock


\bibitem[Giancola et~al\mbox{.}(2022)]%
        {giancola2022soccernet}
\bibfield{author}{\bibinfo{person}{Silvio Giancola}, \bibinfo{person}{Anthony
  Cioppa}, \bibinfo{person}{Adrien Deli{\`e}ge}, \bibinfo{person}{Floriane
  Magera}, \bibinfo{person}{Vladimir Somers}, \bibinfo{person}{Le Kang},
  \bibinfo{person}{Xin Zhou}, \bibinfo{person}{Olivier Barnich},
  \bibinfo{person}{Christophe De~Vleeschouwer}, \bibinfo{person}{Alexandre
  Alahi}, {et~al\mbox{.}}} \bibinfo{year}{2022}\natexlab{}.
\newblock \showarticletitle{SoccerNet 2022 challenges results}. In
  \bibinfo{booktitle}{\emph{Proceedings of the 5th International ACM Workshop
  on Multimedia Content Analysis in Sports}}. \bibinfo{pages}{75--86}.
\newblock


\bibitem[Gou et~al\mbox{.}(2018)]%
        {gou2018systematic}
\bibfield{author}{\bibinfo{person}{Mengran Gou}, \bibinfo{person}{Ziyan Wu},
  \bibinfo{person}{Angels Rates-Borras}, \bibinfo{person}{Octavia Camps},
  \bibinfo{person}{Richard~J Radke}, {et~al\mbox{.}}}
  \bibinfo{year}{2018}\natexlab{}.
\newblock \showarticletitle{A systematic evaluation and benchmark for person
  re-identification: Features, metrics, and datasets}.
\newblock \bibinfo{journal}{\emph{IEEE transactions on pattern analysis and
  machine intelligence}} \bibinfo{volume}{41}, \bibinfo{number}{3}
  (\bibinfo{year}{2018}), \bibinfo{pages}{523--536}.
\newblock


\bibitem[He et~al\mbox{.}(2016)]%
        {he2016deep}
\bibfield{author}{\bibinfo{person}{Kaiming He}, \bibinfo{person}{Xiangyu
  Zhang}, \bibinfo{person}{Shaoqing Ren}, {and} \bibinfo{person}{Jian Sun}.}
  \bibinfo{year}{2016}\natexlab{}.
\newblock \showarticletitle{Deep residual learning for image recognition}. In
  \bibinfo{booktitle}{\emph{Proceedings of the IEEE conference on computer
  vision and pattern recognition}}. \bibinfo{pages}{770--778}.
\newblock


\bibitem[He et~al\mbox{.}(2021)]%
        {he2021transreid}
\bibfield{author}{\bibinfo{person}{Shuting He}, \bibinfo{person}{Hao Luo},
  \bibinfo{person}{Pichao Wang}, \bibinfo{person}{Fan Wang},
  \bibinfo{person}{Hao Li}, {and} \bibinfo{person}{Wei Jiang}.}
  \bibinfo{year}{2021}\natexlab{}.
\newblock \showarticletitle{Transreid: Transformer-based object
  re-identification}. In \bibinfo{booktitle}{\emph{Proceedings of the IEEE/CVF
  international conference on computer vision}}. \bibinfo{pages}{15013--15022}.
\newblock


\bibitem[Hernández-Carrascosa et~al\mbox{.}(2021)]%
        {TGCRBNW}
\bibfield{author}{\bibinfo{person}{Pablo Hernández-Carrascosa},
  \bibinfo{person}{Adrian Penate-Sanchez}, \bibinfo{person}{Javier
  Lorenzo-Navarro}, \bibinfo{person}{David Freire-Obregón}, {and}
  \bibinfo{person}{Modesto Castrillón-Santana}.}
  \bibinfo{year}{2021}\natexlab{}.
\newblock \showarticletitle{TGCRBNW: A Dataset for Runner Bib Number Detection
  (and Recognition) in the Wild}. In \bibinfo{booktitle}{\emph{2020 25th
  International Conference on Pattern Recognition (ICPR)}}.
  \bibinfo{pages}{9445--9451}.
\newblock
\urldef\tempurl%
\url{https://doi.org/10.1109/ICPR48806.2021.9412220}
\showDOI{\tempurl}


\bibitem[Huang et~al\mbox{.}(2018)]%
        {huang2018adversarially}
\bibfield{author}{\bibinfo{person}{Houjing Huang}, \bibinfo{person}{Dangwei
  Li}, \bibinfo{person}{Zhang Zhang}, \bibinfo{person}{Xiaotang Chen}, {and}
  \bibinfo{person}{Kaiqi Huang}.} \bibinfo{year}{2018}\natexlab{}.
\newblock \showarticletitle{Adversarially occluded samples for person
  re-identification}. In \bibinfo{booktitle}{\emph{Proceedings of the IEEE
  conference on computer vision and pattern recognition}}.
  \bibinfo{pages}{5098--5107}.
\newblock


\bibitem[Huang et~al\mbox{.}(2022)]%
        {huang2022swintextspotter}
\bibfield{author}{\bibinfo{person}{Mingxin Huang}, \bibinfo{person}{Yuliang
  Liu}, \bibinfo{person}{Zhenghao Peng}, \bibinfo{person}{Chongyu Liu},
  \bibinfo{person}{Dahua Lin}, \bibinfo{person}{Shenggao Zhu},
  \bibinfo{person}{Nicholas Yuan}, \bibinfo{person}{Kai Ding}, {and}
  \bibinfo{person}{Lianwen Jin}.} \bibinfo{year}{2022}\natexlab{}.
\newblock \showarticletitle{Swintextspotter: Scene text spotting via better
  synergy between text detection and text recognition}. In
  \bibinfo{booktitle}{\emph{proceedings of the IEEE/CVF conference on computer
  vision and pattern recognition}}. \bibinfo{pages}{4593--4603}.
\newblock


\bibitem[Kamlesh et~al\mbox{.}(2017)]%
        {kamlesh2017person}
\bibfield{author}{\bibinfo{person}{Kamlesh}, \bibinfo{person}{Pei Xu},
  \bibinfo{person}{Yang Yang}, {and} \bibinfo{person}{Yongchao Xu}.}
  \bibinfo{year}{2017}\natexlab{}.
\newblock \showarticletitle{Person re-identification with end-to-end scene text
  recognition}. In \bibinfo{booktitle}{\emph{Computer Vision: Second CCF
  Chinese Conference, CCCV 2017, Tianjin, China, October 11--14, 2017,
  Proceedings, Part III}}. Springer, \bibinfo{pages}{363--374}.
\newblock


\bibitem[Karatzas et~al\mbox{.}(2013)]%
        {karatzas2013icdar}
\bibfield{author}{\bibinfo{person}{Dimosthenis Karatzas},
  \bibinfo{person}{Faisal Shafait}, \bibinfo{person}{Seiichi Uchida},
  \bibinfo{person}{Masakazu Iwamura}, \bibinfo{person}{Lluis~Gomez i Bigorda},
  \bibinfo{person}{Sergi~Robles Mestre}, \bibinfo{person}{Joan Mas},
  \bibinfo{person}{David~Fernandez Mota}, \bibinfo{person}{Jon~Almazan
  Almazan}, {and} \bibinfo{person}{Lluis~Pere De~Las~Heras}.}
  \bibinfo{year}{2013}\natexlab{}.
\newblock \showarticletitle{ICDAR 2013 robust reading competition}. In
  \bibinfo{booktitle}{\emph{2013 12th international conference on document
  analysis and recognition}}. IEEE, \bibinfo{pages}{1484--1493}.
\newblock


\bibitem[Krylov et~al\mbox{.}(2021)]%
        {krylov2021open}
\bibfield{author}{\bibinfo{person}{Ilya Krylov}, \bibinfo{person}{Sergei
  Nosov}, {and} \bibinfo{person}{Vladislav Sovrasov}.}
  \bibinfo{year}{2021}\natexlab{}.
\newblock \showarticletitle{Open images v5 text annotation and yet another mask
  text spotter}. In \bibinfo{booktitle}{\emph{Asian Conference on Machine
  Learning}}. PMLR, \bibinfo{pages}{379--389}.
\newblock


\bibitem[Kulkarni et~al\mbox{.}(2018)]%
        {kulkarni2018automatic}
\bibfield{author}{\bibinfo{person}{Yogiraj Kulkarni},
  \bibinfo{person}{Shubhangi Bodkhe}, \bibinfo{person}{Amit Kamthe}, {and}
  \bibinfo{person}{Archana Patil}.} \bibinfo{year}{2018}\natexlab{}.
\newblock \showarticletitle{Automatic number plate recognition for
  motorcyclists riding without helmet}. In \bibinfo{booktitle}{\emph{2018
  International Conference on Current Trends towards Converging Technologies
  (ICCTCT)}}. IEEE, \bibinfo{pages}{1--6}.
\newblock


\bibitem[Laroca et~al\mbox{.}(2018)]%
        {laroca2018robust}
\bibfield{author}{\bibinfo{person}{Rayson Laroca}, \bibinfo{person}{Evair
  Severo}, \bibinfo{person}{Luiz~A Zanlorensi}, \bibinfo{person}{Luiz~S
  Oliveira}, \bibinfo{person}{Gabriel~Resende Gon{\c{c}}alves},
  \bibinfo{person}{William~Robson Schwartz}, {and} \bibinfo{person}{David
  Menotti}.} \bibinfo{year}{2018}\natexlab{}.
\newblock \showarticletitle{A robust real-time automatic license plate
  recognition based on the YOLO detector}. In \bibinfo{booktitle}{\emph{2018
  international joint conference on neural networks (ijcnn)}}. IEEE,
  \bibinfo{pages}{1--10}.
\newblock


\bibitem[Laroca et~al\mbox{.}(2021)]%
        {laroca2021efficient}
\bibfield{author}{\bibinfo{person}{Rayson Laroca}, \bibinfo{person}{Luiz~A
  Zanlorensi}, \bibinfo{person}{Gabriel~R Gon{\c{c}}alves},
  \bibinfo{person}{Eduardo Todt}, \bibinfo{person}{William~Robson Schwartz},
  {and} \bibinfo{person}{David Menotti}.} \bibinfo{year}{2021}\natexlab{}.
\newblock \showarticletitle{An efficient and layout-independent automatic
  license plate recognition system based on the YOLO detector}.
\newblock \bibinfo{journal}{\emph{IET Intelligent Transport Systems}}
  \bibinfo{volume}{15}, \bibinfo{number}{4} (\bibinfo{year}{2021}),
  \bibinfo{pages}{483--503}.
\newblock


\bibitem[Lee et~al\mbox{.}(2004)]%
        {lee2004extraction}
\bibfield{author}{\bibinfo{person}{Hsi-Jian Lee}, \bibinfo{person}{Si-Yuan
  Chen}, {and} \bibinfo{person}{Shen-Zheng Wang}.}
  \bibinfo{year}{2004}\natexlab{}.
\newblock \showarticletitle{Extraction and recognition of license plates of
  motorcycles and vehicles on highways}. In
  \bibinfo{booktitle}{\emph{Proceedings of the 17th International Conference on
  Pattern Recognition, 2004. ICPR 2004.}}, Vol.~\bibinfo{volume}{4}. IEEE,
  \bibinfo{pages}{356--359}.
\newblock


\bibitem[Lee et~al\mbox{.}(2019)]%
        {lee2019snider}
\bibfield{author}{\bibinfo{person}{Younkwan Lee}, \bibinfo{person}{Juhyun Lee},
  \bibinfo{person}{Hoyeon Ahn}, {and} \bibinfo{person}{Moongu Jeon}.}
  \bibinfo{year}{2019}\natexlab{}.
\newblock \showarticletitle{SNIDER: Single noisy image denoising and
  rectification for improving license plate recognition}. In
  \bibinfo{booktitle}{\emph{Proceedings of the IEEE/CVF International
  Conference on Computer Vision Workshops}}. \bibinfo{pages}{0--0}.
\newblock


\bibitem[Li and Liu(2022)]%
        {li2022rider}
\bibfield{author}{\bibinfo{person}{Jiaze Li} {and} \bibinfo{person}{Bin Liu}.}
  \bibinfo{year}{2022}\natexlab{}.
\newblock \showarticletitle{Rider Re-identification Based on Pyramid
  Attention}. In \bibinfo{booktitle}{\emph{Chinese Conference on Pattern
  Recognition and Computer Vision (PRCV)}}. Springer, \bibinfo{pages}{81--93}.
\newblock


\bibitem[Li et~al\mbox{.}(2019)]%
        {li2019global}
\bibfield{author}{\bibinfo{person}{Jianing Li}, \bibinfo{person}{Jingdong
  Wang}, \bibinfo{person}{Qi Tian}, \bibinfo{person}{Wen Gao}, {and}
  \bibinfo{person}{Shiliang Zhang}.} \bibinfo{year}{2019}\natexlab{}.
\newblock \showarticletitle{Global-local temporal representations for video
  person re-identification}. In \bibinfo{booktitle}{\emph{Proceedings of the
  IEEE/CVF international conference on computer vision}}.
  \bibinfo{pages}{3958--3967}.
\newblock


\bibitem[Lin et~al\mbox{.}(2014)]%
        {lin2014microsoft}
\bibfield{author}{\bibinfo{person}{Tsung-Yi Lin}, \bibinfo{person}{Michael
  Maire}, \bibinfo{person}{Serge Belongie}, \bibinfo{person}{James Hays},
  \bibinfo{person}{Pietro Perona}, \bibinfo{person}{Deva Ramanan},
  \bibinfo{person}{Piotr Doll{\'a}r}, {and} \bibinfo{person}{C~Lawrence
  Zitnick}.} \bibinfo{year}{2014}\natexlab{}.
\newblock \showarticletitle{Microsoft coco: Common objects in context}. In
  \bibinfo{booktitle}{\emph{Computer Vision--ECCV 2014: 13th European
  Conference, Zurich, Switzerland, September 6-12, 2014, Proceedings, Part V
  13}}. Springer, \bibinfo{pages}{740--755}.
\newblock


\bibitem[Liu and Bhanu(2019)]%
        {liu2019pose}
\bibfield{author}{\bibinfo{person}{Hengyue Liu} {and} \bibinfo{person}{Bir
  Bhanu}.} \bibinfo{year}{2019}\natexlab{}.
\newblock \showarticletitle{Pose-guided R-CNN for jersey number recognition in
  sports}. In \bibinfo{booktitle}{\emph{Proceedings of the IEEE/CVF Conference
  on Computer Vision and Pattern Recognition Workshops}}.
  \bibinfo{pages}{0--0}.
\newblock


\bibitem[Lyu et~al\mbox{.}(2018)]%
        {lyu2018mask}
\bibfield{author}{\bibinfo{person}{Pengyuan Lyu}, \bibinfo{person}{Minghui
  Liao}, \bibinfo{person}{Cong Yao}, \bibinfo{person}{Wenhao Wu}, {and}
  \bibinfo{person}{Xiang Bai}.} \bibinfo{year}{2018}\natexlab{}.
\newblock \showarticletitle{Mask textspotter: An end-to-end trainable neural
  network for spotting text with arbitrary shapes}. In
  \bibinfo{booktitle}{\emph{Proceedings of the European conference on computer
  vision (ECCV)}}. \bibinfo{pages}{67--83}.
\newblock


\bibitem[Messelodi and Modena(2013)]%
        {messelodi2013scene}
\bibfield{author}{\bibinfo{person}{Stefano Messelodi} {and}
  \bibinfo{person}{Carla~Maria Modena}.} \bibinfo{year}{2013}\natexlab{}.
\newblock \showarticletitle{Scene text recognition and tracking to identify
  athletes in sport videos}.
\newblock \bibinfo{journal}{\emph{Multimedia tools and applications}}
  \bibinfo{volume}{63}, \bibinfo{number}{2} (\bibinfo{year}{2013}),
  \bibinfo{pages}{521--545}.
\newblock


\bibitem[Nag et~al\mbox{.}(2019)]%
        {nag2019crnn}
\bibfield{author}{\bibinfo{person}{Sauradip Nag}, \bibinfo{person}{Raghavendra
  Ramachandra}, \bibinfo{person}{Palaiahnakote Shivakumara},
  \bibinfo{person}{Umapada Pal}, \bibinfo{person}{Tong Lu}, {and}
  \bibinfo{person}{Mohan Kankanhalli}.} \bibinfo{year}{2019}\natexlab{}.
\newblock \showarticletitle{CRNN based jersey-bib number/text recognition in
  sports and marathon images}. In \bibinfo{booktitle}{\emph{2019 International
  Conference on Document Analysis and Recognition (ICDAR)}}. IEEE,
  \bibinfo{pages}{1149--1156}.
\newblock


\bibitem[Netzer et~al\mbox{.}(2011)]%
        {netzer2011reading}
\bibfield{author}{\bibinfo{person}{Yuval Netzer}, \bibinfo{person}{Tao Wang},
  \bibinfo{person}{Adam Coates}, \bibinfo{person}{Alessandro Bissacco},
  \bibinfo{person}{Bo Wu}, {and} \bibinfo{person}{Andrew~Y Ng}.}
  \bibinfo{year}{2011}\natexlab{}.
\newblock \showarticletitle{Reading digits in natural images with unsupervised
  feature learning}.
\newblock  (\bibinfo{year}{2011}).
\newblock


\bibitem[Penate-Sanchez et~al\mbox{.}(2020)]%
        {PENATESANCHEZ2020355}
\bibfield{author}{\bibinfo{person}{Adrian Penate-Sanchez},
  \bibinfo{person}{David Freire-Obregón}, \bibinfo{person}{Adrián
  Lorenzo-Melián}, \bibinfo{person}{Javier Lorenzo-Navarro}, {and}
  \bibinfo{person}{Modesto Castrillón-Santana}.}
  \bibinfo{year}{2020}\natexlab{}.
\newblock \showarticletitle{TGC20ReId: A dataset for sport event
  re-identification in the wild}.
\newblock \bibinfo{journal}{\emph{Pattern Recognition Letters}}
  \bibinfo{volume}{138} (\bibinfo{year}{2020}), \bibinfo{pages}{355--361}.
\newblock
\showISSN{0167-8655}
\urldef\tempurl%
\url{https://doi.org/10.1016/j.patrec.2020.08.003}
\showDOI{\tempurl}


\bibitem[Quang et~al\mbox{.}(2022)]%
        {quang2022character}
\bibfield{author}{\bibinfo{person}{Huy~Che Quang}, \bibinfo{person}{Tung
  Do~Thanh}, {and} \bibinfo{person}{Cuong~Truong Van}.}
  \bibinfo{year}{2022}\natexlab{}.
\newblock \showarticletitle{Character Time-series Matching For Robust License
  Plate Recognition}. In \bibinfo{booktitle}{\emph{2022 International
  Conference on Multimedia Analysis and Pattern Recognition (MAPR)}}. IEEE,
  \bibinfo{pages}{1--6}.
\newblock


\bibitem[Ristani et~al\mbox{.}(2016)]%
        {ristani2016performance}
\bibfield{author}{\bibinfo{person}{Ergys Ristani}, \bibinfo{person}{Francesco
  Solera}, \bibinfo{person}{Roger Zou}, \bibinfo{person}{Rita Cucchiara}, {and}
  \bibinfo{person}{Carlo Tomasi}.} \bibinfo{year}{2016}\natexlab{}.
\newblock \showarticletitle{Performance measures and a data set for
  multi-target, multi-camera tracking}. In \bibinfo{booktitle}{\emph{European
  conference on computer vision}}. Springer, \bibinfo{pages}{17--35}.
\newblock


\bibitem[Sanjana et~al\mbox{.}(2021)]%
        {sanjana2021review}
\bibfield{author}{\bibinfo{person}{S Sanjana}, \bibinfo{person}{S Sanjana},
  \bibinfo{person}{VR Shriya}, \bibinfo{person}{Gururaj Vaishnavi}, {and}
  \bibinfo{person}{K Ashwini}.} \bibinfo{year}{2021}\natexlab{}.
\newblock \showarticletitle{A review on various methodologies used for vehicle
  classification, helmet detection and number plate recognition}.
\newblock \bibinfo{journal}{\emph{Evolutionary Intelligence}}
  \bibinfo{volume}{14}, \bibinfo{number}{2} (\bibinfo{year}{2021}),
  \bibinfo{pages}{979--987}.
\newblock


\bibitem[{\v{S}}aric et~al\mbox{.}(2008)]%
        {vsaric2008player}
\bibfield{author}{\bibinfo{person}{Matko {\v{S}}aric}, \bibinfo{person}{Hrvoje
  Dujmic}, \bibinfo{person}{Vladan Papic}, {and} \bibinfo{person}{Nikola
  Ro{\v{z}}ic}.} \bibinfo{year}{2008}\natexlab{}.
\newblock \showarticletitle{Player number localization and recognition in
  soccer video using hsv color space and internal contours}.
\newblock \bibinfo{journal}{\emph{International Journal of Electrical and
  Computer Engineering}} \bibinfo{volume}{2}, \bibinfo{number}{7}
  (\bibinfo{year}{2008}), \bibinfo{pages}{1408--1412}.
\newblock


\bibitem[Sathe et~al\mbox{.}(2022)]%
        {sathe2022helmet}
\bibfield{author}{\bibinfo{person}{Pushkar Sathe}, \bibinfo{person}{Aditi Rao},
  \bibinfo{person}{Aditya Singh}, \bibinfo{person}{Ritika Nair}, {and}
  \bibinfo{person}{Abhilash Poojary}.} \bibinfo{year}{2022}\natexlab{}.
\newblock \showarticletitle{Helmet Detection And Number Plate Recognition Using
  Deep Learning}. In \bibinfo{booktitle}{\emph{2022 IEEE Region 10 Symposium
  (TENSYMP)}}. IEEE, \bibinfo{pages}{1--6}.
\newblock


\bibitem[Schwartz and Davis(2009)]%
        {schwartz2009learning}
\bibfield{author}{\bibinfo{person}{William~Robson Schwartz} {and}
  \bibinfo{person}{Larry~S Davis}.} \bibinfo{year}{2009}\natexlab{}.
\newblock \showarticletitle{Learning discriminative appearance-based models
  using partial least squares}. In \bibinfo{booktitle}{\emph{2009 XXII
  Brazilian symposium on computer graphics and image processing}}. IEEE,
  \bibinfo{pages}{322--329}.
\newblock


\bibitem[Shashirangana et~al\mbox{.}(2020)]%
        {shashirangana2020automated}
\bibfield{author}{\bibinfo{person}{Jithmi Shashirangana},
  \bibinfo{person}{Heshan Padmasiri}, \bibinfo{person}{Dulani Meedeniya}, {and}
  \bibinfo{person}{Charith Perera}.} \bibinfo{year}{2020}\natexlab{}.
\newblock \showarticletitle{Automated license plate recognition: a survey on
  methods and techniques}.
\newblock \bibinfo{journal}{\emph{IEEE Access}}  \bibinfo{volume}{9}
  (\bibinfo{year}{2020}), \bibinfo{pages}{11203--11225}.
\newblock


\bibitem[Shi et~al\mbox{.}(2019)]%
        {aster}
\bibfield{author}{\bibinfo{person}{Baoguang Shi}, \bibinfo{person}{Mingkun
  Yang}, \bibinfo{person}{Xinggang Wang}, \bibinfo{person}{Pengyuan Lyu},
  \bibinfo{person}{Cong Yao}, {and} \bibinfo{person}{Xiang Bai}.}
  \bibinfo{year}{2019}\natexlab{}.
\newblock \showarticletitle{ASTER: An Attentional Scene Text Recognizer with
  Flexible Rectification}.
\newblock \bibinfo{journal}{\emph{IEEE Transactions on Pattern Analysis and
  Machine Intelligence}} \bibinfo{volume}{41}, \bibinfo{number}{9}
  (\bibinfo{year}{2019}), \bibinfo{pages}{2035--2048}.
\newblock
\urldef\tempurl%
\url{https://doi.org/10.1109/TPAMI.2018.2848939}
\showDOI{\tempurl}


\bibitem[Shi et~al\mbox{.}(2015)]%
        {shi2015transferring}
\bibfield{author}{\bibinfo{person}{Zhiyuan Shi}, \bibinfo{person}{Timothy~M
  Hospedales}, {and} \bibinfo{person}{Tao Xiang}.}
  \bibinfo{year}{2015}\natexlab{}.
\newblock \showarticletitle{Transferring a semantic representation for person
  re-identification and search}. In \bibinfo{booktitle}{\emph{Proceedings of
  the IEEE Conference on Computer Vision and Pattern Recognition}}.
  \bibinfo{pages}{4184--4193}.
\newblock


\bibitem[Shivakumara et~al\mbox{.}(2017)]%
        {shivakumara2017new}
\bibfield{author}{\bibinfo{person}{Palaiahnakote Shivakumara},
  \bibinfo{person}{Ramachandra Raghavendra}, \bibinfo{person}{Longfei Qin},
  \bibinfo{person}{Kiran~B Raja}, \bibinfo{person}{Tong Lu}, {and}
  \bibinfo{person}{Umapada Pal}.} \bibinfo{year}{2017}\natexlab{}.
\newblock \showarticletitle{A new multi-modal approach to bib number/text
  detection and recognition in Marathon images}.
\newblock \bibinfo{journal}{\emph{pattern recognition}}  \bibinfo{volume}{61}
  (\bibinfo{year}{2017}), \bibinfo{pages}{479--491}.
\newblock


\bibitem[Silva and Jung(2018)]%
        {silva2018license}
\bibfield{author}{\bibinfo{person}{Sergio~Montazzolli Silva} {and}
  \bibinfo{person}{Claudio~Rosito Jung}.} \bibinfo{year}{2018}\natexlab{}.
\newblock \showarticletitle{License plate detection and recognition in
  unconstrained scenarios}. In \bibinfo{booktitle}{\emph{Proceedings of the
  European conference on computer vision (ECCV)}}. \bibinfo{pages}{580--596}.
\newblock


\bibitem[Tian et~al\mbox{.}(2016)]%
        {tian2016detecting}
\bibfield{author}{\bibinfo{person}{Zhi Tian}, \bibinfo{person}{Weilin Huang},
  \bibinfo{person}{Tong He}, \bibinfo{person}{Pan He}, {and}
  \bibinfo{person}{Yu Qiao}.} \bibinfo{year}{2016}\natexlab{}.
\newblock \showarticletitle{Detecting text in natural image with connectionist
  text proposal network}. In \bibinfo{booktitle}{\emph{Computer Vision--ECCV
  2016: 14th European Conference, Amsterdam, The Netherlands, October 11-14,
  2016, Proceedings, Part VIII 14}}. Springer, \bibinfo{pages}{56--72}.
\newblock


\bibitem[Van~Zandycke et~al\mbox{.}(2022)]%
        {van2022deepsportradar}
\bibfield{author}{\bibinfo{person}{Gabriel Van~Zandycke},
  \bibinfo{person}{Vladimir Somers}, \bibinfo{person}{Maxime Istasse},
  \bibinfo{person}{Carlo~Del Don}, {and} \bibinfo{person}{Davide Zambrano}.}
  \bibinfo{year}{2022}\natexlab{}.
\newblock \showarticletitle{Deepsportradar-v1: Computer vision dataset for
  sports understanding with high quality annotations}. In
  \bibinfo{booktitle}{\emph{Proceedings of the 5th International ACM Workshop
  on Multimedia Content Analysis in Sports}}. \bibinfo{pages}{1--8}.
\newblock


\bibitem[Vats et~al\mbox{.}(2021)]%
        {vats2021multi}
\bibfield{author}{\bibinfo{person}{Kanav Vats}, \bibinfo{person}{Mehrnaz Fani},
  \bibinfo{person}{David~A Clausi}, {and} \bibinfo{person}{John Zelek}.}
  \bibinfo{year}{2021}\natexlab{}.
\newblock \showarticletitle{Multi-task learning for jersey number recognition
  in ice hockey}. In \bibinfo{booktitle}{\emph{Proceedings of the 4th
  International Workshop on Multimedia Content Analysis in Sports}}.
  \bibinfo{pages}{11--15}.
\newblock


\bibitem[Veit et~al\mbox{.}(2016)]%
        {veit2016coco}
\bibfield{author}{\bibinfo{person}{Andreas Veit}, \bibinfo{person}{Tomas
  Matera}, \bibinfo{person}{Lukas Neumann}, \bibinfo{person}{Jiri Matas}, {and}
  \bibinfo{person}{Serge Belongie}.} \bibinfo{year}{2016}\natexlab{}.
\newblock \showarticletitle{Coco-text: Dataset and benchmark for text detection
  and recognition in natural images}.
\newblock \bibinfo{journal}{\emph{arXiv preprint arXiv:1601.07140}}
  (\bibinfo{year}{2016}).
\newblock


\bibitem[Wang et~al\mbox{.}(2011a)]%
        {wang2011end}
\bibfield{author}{\bibinfo{person}{Kai Wang}, \bibinfo{person}{Boris Babenko},
  {and} \bibinfo{person}{Serge Belongie}.} \bibinfo{year}{2011}\natexlab{a}.
\newblock \showarticletitle{End-to-end scene text recognition}. In
  \bibinfo{booktitle}{\emph{2011 International conference on computer vision}}.
  IEEE, \bibinfo{pages}{1457--1464}.
\newblock


\bibitem[Wang and Belongie(2010)]%
        {wang2010word}
\bibfield{author}{\bibinfo{person}{Kai Wang} {and} \bibinfo{person}{Serge
  Belongie}.} \bibinfo{year}{2010}\natexlab{}.
\newblock \showarticletitle{Word spotting in the wild}. In
  \bibinfo{booktitle}{\emph{Computer Vision--ECCV 2010: 11th European
  Conference on Computer Vision, Heraklion, Crete, Greece, September 5-11,
  2010, Proceedings, Part I 11}}. Springer, \bibinfo{pages}{591--604}.
\newblock


\bibitem[Wang et~al\mbox{.}(2019)]%
        {CRNN}
\bibfield{author}{\bibinfo{person}{Ruishuang Wang}, \bibinfo{person}{Zhao Li},
  \bibinfo{person}{Jian Cao}, \bibinfo{person}{Tong Chen}, {and}
  \bibinfo{person}{Lei Wang}.} \bibinfo{year}{2019}\natexlab{}.
\newblock \showarticletitle{Convolutional Recurrent Neural Networks for Text
  Classification}. In \bibinfo{booktitle}{\emph{2019 International Joint
  Conference on Neural Networks (IJCNN)}}. \bibinfo{pages}{1--6}.
\newblock
\urldef\tempurl%
\url{https://doi.org/10.1109/IJCNN.2019.8852406}
\showDOI{\tempurl}


\bibitem[Wang et~al\mbox{.}(2011b)]%
        {wang2011re}
\bibfield{author}{\bibinfo{person}{Simi Wang}, \bibinfo{person}{Michal
  Lewandowski}, \bibinfo{person}{James Annesley}, {and} \bibinfo{person}{James
  Orwell}.} \bibinfo{year}{2011}\natexlab{b}.
\newblock \showarticletitle{Re-identification of pedestrians with variable
  occlusion and scale}. In \bibinfo{booktitle}{\emph{2011 IEEE International
  Conference on Computer Vision Workshops (ICCV Workshops)}}. IEEE,
  \bibinfo{pages}{1876--1882}.
\newblock


\bibitem[Wang et~al\mbox{.}(2016)]%
        {wang2016person}
\bibfield{author}{\bibinfo{person}{Taiqing Wang}, \bibinfo{person}{Shaogang
  Gong}, \bibinfo{person}{Xiatian Zhu}, {and} \bibinfo{person}{Shengjin Wang}.}
  \bibinfo{year}{2016}\natexlab{}.
\newblock \showarticletitle{Person re-identification by discriminative
  selection in video ranking}.
\newblock \bibinfo{journal}{\emph{IEEE transactions on pattern analysis and
  machine intelligence}} \bibinfo{volume}{38}, \bibinfo{number}{12}
  (\bibinfo{year}{2016}), \bibinfo{pages}{2501--2514}.
\newblock


\bibitem[Wei et~al\mbox{.}(2018)]%
        {wei2018person}
\bibfield{author}{\bibinfo{person}{Longhui Wei}, \bibinfo{person}{Shiliang
  Zhang}, \bibinfo{person}{Wen Gao}, {and} \bibinfo{person}{Qi Tian}.}
  \bibinfo{year}{2018}\natexlab{}.
\newblock \showarticletitle{Person transfer gan to bridge domain gap for person
  re-identification}. In \bibinfo{booktitle}{\emph{Proceedings of the IEEE
  conference on computer vision and pattern recognition}}.
  \bibinfo{pages}{79--88}.
\newblock


\bibitem[Wro{\'n}ska et~al\mbox{.}(2017)]%
        {wronska2017athlete}
\bibfield{author}{\bibinfo{person}{Ada Wro{\'n}ska}, \bibinfo{person}{Kacper
  Sarnacki}, {and} \bibinfo{person}{Khalid Saeed}.}
  \bibinfo{year}{2017}\natexlab{}.
\newblock \showarticletitle{Athlete number detection on the basis of their face
  images}. In \bibinfo{booktitle}{\emph{2017 International Conference on
  Biometrics and Kansei Engineering (ICBAKE)}}. IEEE, \bibinfo{pages}{84--89}.
\newblock


\bibitem[Xiao et~al\mbox{.}(2016)]%
        {xiao2016end}
\bibfield{author}{\bibinfo{person}{Tong Xiao}, \bibinfo{person}{Shuang Li},
  \bibinfo{person}{Bochao Wang}, \bibinfo{person}{Liang Lin}, {and}
  \bibinfo{person}{Xiaogang Wang}.} \bibinfo{year}{2016}\natexlab{}.
\newblock \showarticletitle{End-to-end deep learning for person search}.
\newblock \bibinfo{journal}{\emph{arXiv preprint arXiv:1604.01850}}
  \bibinfo{volume}{2}, \bibinfo{number}{2} (\bibinfo{year}{2016}),
  \bibinfo{pages}{4}.
\newblock


\bibitem[Yamamoto et~al\mbox{.}(2013)]%
        {yamamoto2013multiple}
\bibfield{author}{\bibinfo{person}{Taiki Yamamoto}, \bibinfo{person}{Hirokatsu
  Kataoka}, \bibinfo{person}{Masaki Hayashi}, \bibinfo{person}{Yoshimitsu
  Aoki}, \bibinfo{person}{Kyoko Oshima}, {and} \bibinfo{person}{Masamoto
  Tanabiki}.} \bibinfo{year}{2013}\natexlab{}.
\newblock \showarticletitle{Multiple players tracking and identification using
  group detection and player number recognition in sports video}. In
  \bibinfo{booktitle}{\emph{IECON 2013-39th Annual Conference of the IEEE
  Industrial Electronics Society}}. IEEE, \bibinfo{pages}{2442--2446}.
\newblock


\bibitem[Ye et~al\mbox{.}(2021)]%
        {ye2021deep}
\bibfield{author}{\bibinfo{person}{Mang Ye}, \bibinfo{person}{Jianbing Shen},
  \bibinfo{person}{Gaojie Lin}, \bibinfo{person}{Tao Xiang},
  \bibinfo{person}{Ling Shao}, {and} \bibinfo{person}{Steven~CH Hoi}.}
  \bibinfo{year}{2021}\natexlab{}.
\newblock \showarticletitle{Deep learning for person re-identification: A
  survey and outlook}.
\newblock \bibinfo{journal}{\emph{IEEE transactions on pattern analysis and
  machine intelligence}} \bibinfo{volume}{44}, \bibinfo{number}{6}
  (\bibinfo{year}{2021}), \bibinfo{pages}{2872--2893}.
\newblock


\bibitem[Ye et~al\mbox{.}(2023)]%
        {Ye_2023_CVPR}
\bibfield{author}{\bibinfo{person}{Maoyuan Ye}, \bibinfo{person}{Jing Zhang},
  \bibinfo{person}{Shanshan Zhao}, \bibinfo{person}{Juhua Liu},
  \bibinfo{person}{Tongliang Liu}, \bibinfo{person}{Bo Du}, {and}
  \bibinfo{person}{Dacheng Tao}.} \bibinfo{year}{2023}\natexlab{}.
\newblock \showarticletitle{DeepSolo: Let Transformer Decoder With Explicit
  Points Solo for Text Spotting}. In \bibinfo{booktitle}{\emph{Proceedings of
  the IEEE/CVF Conference on Computer Vision and Pattern Recognition (CVPR)}}.
  \bibinfo{pages}{19348--19357}.
\newblock


\bibitem[Yi et~al\mbox{.}(2014)]%
        {yi2014deep}
\bibfield{author}{\bibinfo{person}{Dong Yi}, \bibinfo{person}{Zhen Lei},
  \bibinfo{person}{Shengcai Liao}, {and} \bibinfo{person}{Stan~Z Li}.}
  \bibinfo{year}{2014}\natexlab{}.
\newblock \showarticletitle{Deep metric learning for person re-identification}.
  In \bibinfo{booktitle}{\emph{2014 22nd international conference on pattern
  recognition}}. IEEE, \bibinfo{pages}{34--39}.
\newblock


\bibitem[Yuan et~al\mbox{.}(2018)]%
        {yuan2018bike}
\bibfield{author}{\bibinfo{person}{Yuan Yuan}, \bibinfo{person}{Jian’an
  Zhang}, {and} \bibinfo{person}{Qi Wang}.} \bibinfo{year}{2018}\natexlab{}.
\newblock \showarticletitle{Bike-person re-identification: a benchmark and a
  comprehensive evaluation}.
\newblock \bibinfo{journal}{\emph{IEEE Access}}  \bibinfo{volume}{6}
  (\bibinfo{year}{2018}), \bibinfo{pages}{56059--56068}.
\newblock


\bibitem[Zhao et~al\mbox{.}(2023)]%
        {zhao2023clip4str}
\bibfield{author}{\bibinfo{person}{Shuai Zhao}, \bibinfo{person}{Xiaohan Wang},
  \bibinfo{person}{Linchao Zhu}, {and} \bibinfo{person}{Yi Yang}.}
  \bibinfo{year}{2023}\natexlab{}.
\newblock \bibinfo{title}{CLIP4STR: A Simple Baseline for Scene Text
  Recognition with Pre-trained Vision-Language Model}.
\newblock
\newblock
\showeprint[arxiv]{2305.14014}~[cs.CV]


\bibitem[Zheng et~al\mbox{.}(2015)]%
        {zheng2015scalable}
\bibfield{author}{\bibinfo{person}{Liang Zheng}, \bibinfo{person}{Liyue Shen},
  \bibinfo{person}{Lu Tian}, \bibinfo{person}{Shengjin Wang},
  \bibinfo{person}{Jingdong Wang}, {and} \bibinfo{person}{Qi Tian}.}
  \bibinfo{year}{2015}\natexlab{}.
\newblock \showarticletitle{Scalable person re-identification: A benchmark}. In
  \bibinfo{booktitle}{\emph{Proceedings of the IEEE international conference on
  computer vision}}. \bibinfo{pages}{1116--1124}.
\newblock


\bibitem[Zheng et~al\mbox{.}(2016)]%
        {zheng2016person}
\bibfield{author}{\bibinfo{person}{Liang Zheng}, \bibinfo{person}{Yi Yang},
  {and} \bibinfo{person}{Alexander~G Hauptmann}.}
  \bibinfo{year}{2016}\natexlab{}.
\newblock \showarticletitle{Person re-identification: Past, present and
  future}.
\newblock \bibinfo{journal}{\emph{arXiv preprint arXiv:1610.02984}}
  (\bibinfo{year}{2016}).
\newblock


\bibitem[Zhou et~al\mbox{.}(2019)]%
        {zhou2019omni}
\bibfield{author}{\bibinfo{person}{Kaiyang Zhou}, \bibinfo{person}{Yongxin
  Yang}, \bibinfo{person}{Andrea Cavallaro}, {and} \bibinfo{person}{Tao
  Xiang}.} \bibinfo{year}{2019}\natexlab{}.
\newblock \showarticletitle{Omni-scale feature learning for person
  re-identification}. In \bibinfo{booktitle}{\emph{Proceedings of the IEEE/CVF
  international conference on computer vision}}. \bibinfo{pages}{3702--3712}.
\newblock


\bibitem[Zhou et~al\mbox{.}(2017)]%
        {Zhou_2017_CVPR}
\bibfield{author}{\bibinfo{person}{Xinyu Zhou}, \bibinfo{person}{Cong Yao},
  \bibinfo{person}{He Wen}, \bibinfo{person}{Yuzhi Wang},
  \bibinfo{person}{Shuchang Zhou}, \bibinfo{person}{Weiran He}, {and}
  \bibinfo{person}{Jiajun Liang}.} \bibinfo{year}{2017}\natexlab{}.
\newblock \showarticletitle{EAST: An Efficient and Accurate Scene Text
  Detector}. In \bibinfo{booktitle}{\emph{Proceedings of the IEEE Conference on
  Computer Vision and Pattern Recognition (CVPR)}}.
\newblock


\end{thebibliography}

\clearpage
\appendix
\section{Hyperparameter Searches}
\label{sec:hyperparameters}

For all experiments in this paper, 
a grid search was used on the training and validation set to 
select the best-performing hyperparameters after 50 epochs. 
Then given the best hyperparameters, 
a final training run was performed on the combined training and validation sets, 
and then the best reported metrics on the test set were reported. 
The hyperparameter search ranges for the RnD dataset are in Table~\ref{tab:rnd-sweep-config} 
and in Table~\ref{tab:mudd-sweep-config} for MUDD. 
The best-performing hyperparameters are shown in bold.

\begin{table}[thb]
    \centering
    \caption{Hyperparmeter sweep ranges for the models trained on the RnD dataset.}
    \label{tab:rnd-sweep-config}
    \begin{tabular}{ll}
    \hline
    Parameter & Configuration \\ \hline
    Learning Rate & [0.0001, \textbf{0.001} 0.01] \\
    Learning Rate Schedule & [linear, \textbf{cosine}, one-cycle] \\
    Warm-Up Period & [0, 100, \textbf{1000}, 2000]\\
    Batch Size & [8, 16, \textbf{32}] \\ \bottomrule
    \end{tabular}
\end{table}

\begin{table}[thb]
    \centering
    \caption{Hyperparmeter sweep ranges for the models trained on the MUDD dataset.}
    \label{tab:mudd-sweep-config}
    \begin{tabular}{ll}
    \hline
    Parameter & Configuration \\ \hline
    Learning Rate & [0.0001, \textbf{0.0003}, 0.001, 0.003, 0.01]\\
    Learning Rate Schedule & [linear, \textbf{cosine}, one-cycle] \\
    Data Augmentation & \begin{tabular}[c]{@{}l@{}}
        [\textbf{random flip, color jitter,} \\ \ \ \textbf{random crop}, resize, \\ \ \ random rotation]
        \end{tabular} \\
    Warm-Up Period & [0, 100, \textbf{1000}, 2000]\\
    Batch Size & [32, 64, \textbf{128}, 256] \\ \bottomrule
    \end{tabular}
\end{table}

\section{Efficient ReID Labeling via Auxiliary Information}
\label{sec:mudd-labeling-info}

\begin{figure}[thb]
    \includegraphics[width=0.5\textwidth]{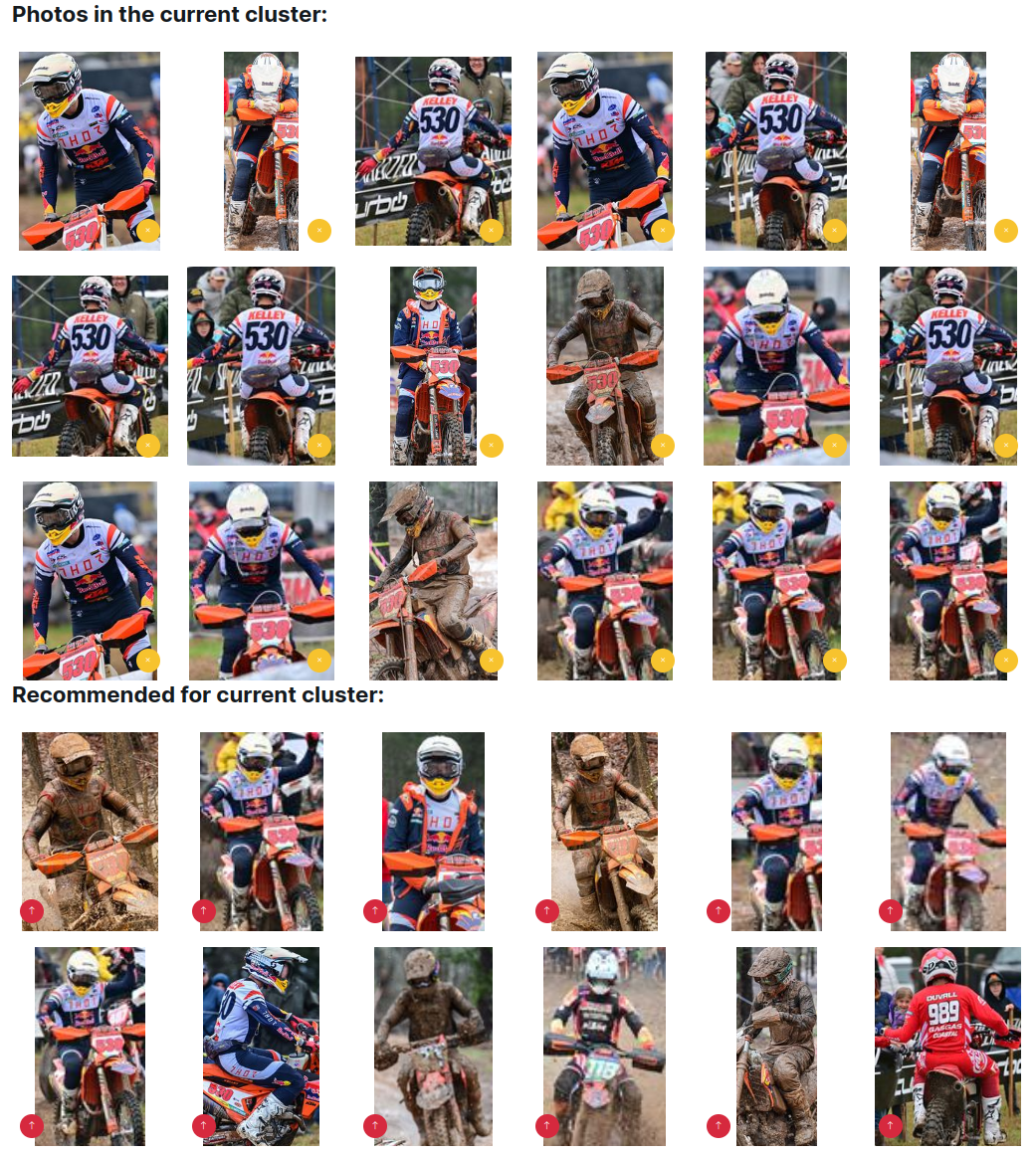}
    \caption{Leveraging detected jersey numbers as auxiliary information enables generating higher quality identity clustering proposals for manual verification. This proposed cluster contains both clean and muddy images of the same rider, whereas proposing clusters with off-the-shelf re-id models fail.}
    \label{fig:label_example1}
\end{figure}

\begin{figure}[thb]
    \centering
    \includegraphics[width=0.5\textwidth]{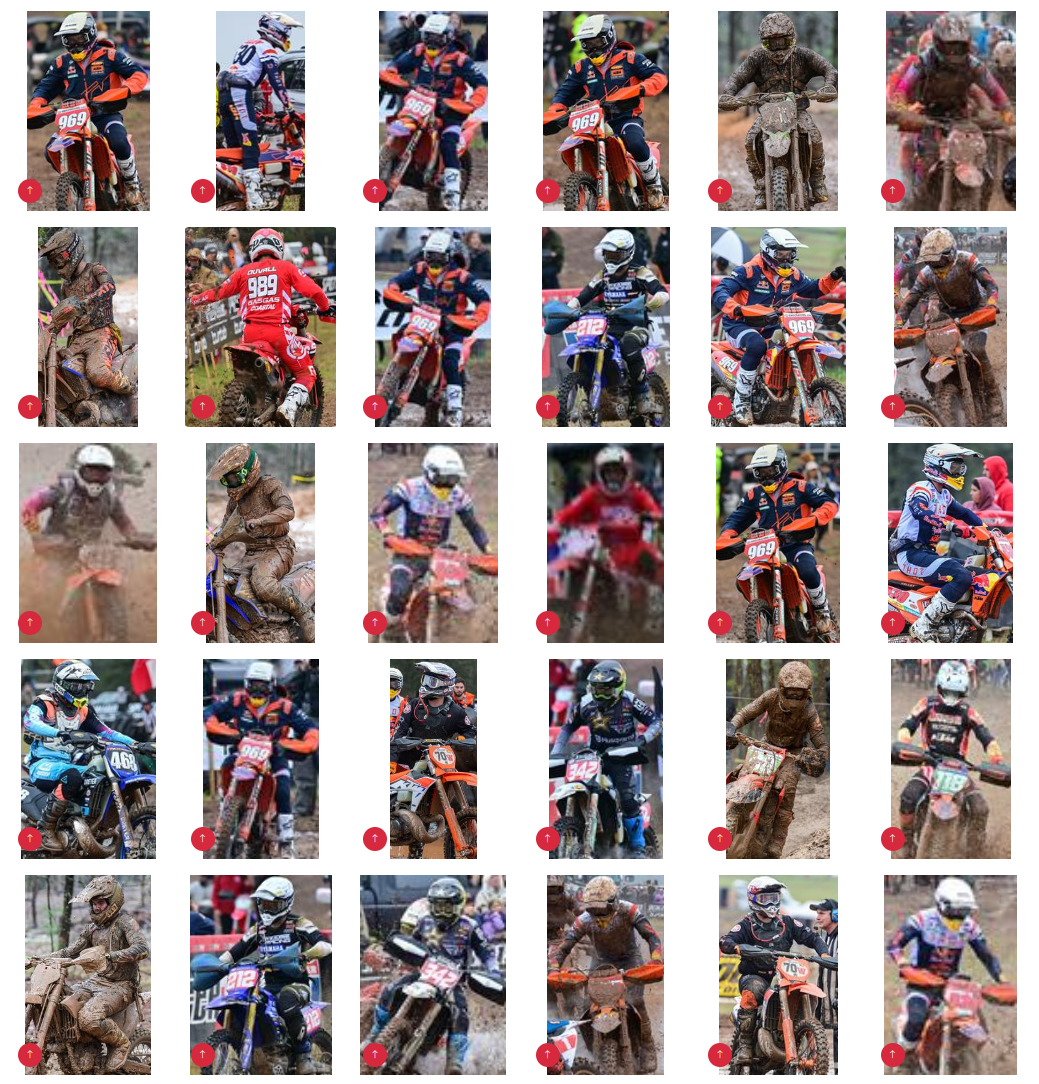}
    \caption{Additional proposed results for the same identity cluster as Figure~\ref{fig:label_example1}. Our methodology provides high-quality recommendations to simplify manual verification and labeling.}
    \label{fig:label_example2}
\end{figure}

One key challenge in constructing re-id datasets 
is to efficiently group images 
of the same identity during labeling. 
Exhaustively labeling identities from scratch can become
intractable for a large dataset of images 
with an unknown number of identities. 
To assist in this labeling process, 
images can be clustered 
into groups using pretrained models,
and then manually verified by annotators. 
However, in constructing MUDD, 
we found that annotators still spent over 30 
minutes on each identity, 
requiring a more efficient process.

Off-the-shelf re-id models focus 
on extracting features invariant 
to nuisance factors like pose, 
lighting, and blurring
while discriminating between identities. 
However, these features are based on their pretraining dataset
and they cannot explicitly 
leverage domain-specific cues---especially
if the domain-specific cues are not available 
in the pretraining dataset, such as racer numbers. 
The re-id model treats 
the image holistically 
without localizing and 
recognizing semantic concepts like digits.
Therefore, when the models are applied 
on different image domains, 
any useful domain-specific cues are not used.

In light of this challenge, 
we leverage the fact that each identity (i.e. racer) 
in this dataset is assigned a visible number 
and we propose directly utilizing this auxiliary information 
during the clustering and re-id process via 
a pretrained text detection model~\citep{lyu2018mask}.
This domain knowledge provides 
strong localization cues 
to group images with the same numbers. 
The re-id model alone struggles 
to consistently spot 
and match the small-digit regions 
amidst mud, motion, and variations.

Explicitly guiding search and clustering with the auxiliary numbers, 
even if noisy, complements the holistic re-id model. 
Our breadth-first attribute search 
leverages the domain knowledge 
to effectively explore the data 
and retrieve number matches. 
This creates high-quality initial clusters 
that seed the depth-first re-id search.

In essence, we get the best of both worlds:
domain-driven localization from the auxiliary cues, 
combined with holistic identity discrimination from the re-id model. 
The re-id model alone lacks 
the explicit semantic guidance, 
resulting in poorer search and clustering. 
Our hybrid approach better utilizes 
both domain knowledge and learned representations.

Specifically,
to generate ground truth labels for specific racers, 
we first extract all numbers using a
pretrained text detection model~\citep{lyu2018mask}, 
and also create a re-id embedding using a pretrained OSNet model.
Then we iterate over the following process:
\begin{enumerate}
    \item Pick a number that was detected more than 10 times and retrieve all images containing it.
    \item For each result from Step 1, 
    take the top $k$ nearest neighbors based on the re-id embedding.
    \item Combine the results for each search by rank, 
    and present 
    to annotators for manual refinement and verification.
\end{enumerate}
This updated process reduced the average time 
to verify an identity cluster 
from over 30 minutes to under 10. 

Figure~\ref{fig:label_example1}
shows a proposed cluster from our labeling system.
The top section contains all photos where the number 
530 was detected. 
The bottom section shows the most similar images
according to the pertained OSNet re-id model. 
Critically, leveraging the auxiliary number information provides 
an initial cluster with clean and muddy images 
of the same racer that can be used as a seed image 
for a search by the re-id model. 
Figure \ref{fig:label_example2} shows additional results deeper in the ranking. 

\section{Additional Text Spotting Results}
\label{sec:additional-ts-results}

\begin{figure*}[thb]
    \centering
    \begin{subfigure}[b]{0.24\textwidth}
        \centering
        \includegraphics[width=\textwidth]{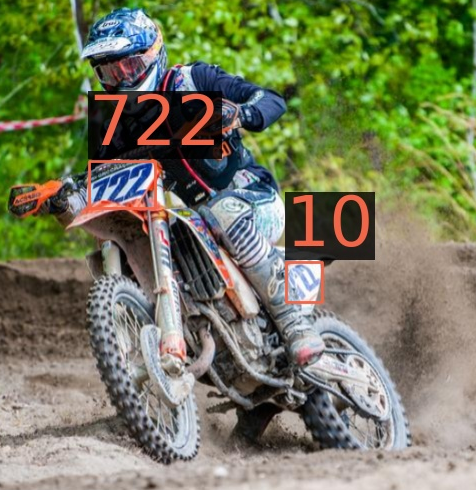}
        \caption{}
        \label{img:backbikefail}
    \end{subfigure}
    \hfill 
    \begin{subfigure}[b]{0.165\textwidth}
        \centering
        \includegraphics[width=\textwidth]{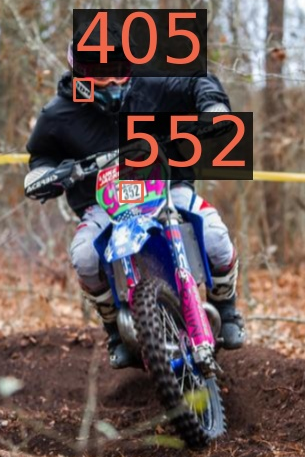}
        \caption{}
        \label{img:stackednums}
    \end{subfigure}
    \hfill
    \begin{subfigure}[b]{0.162\textwidth}
        \centering
        \includegraphics[width=\textwidth]{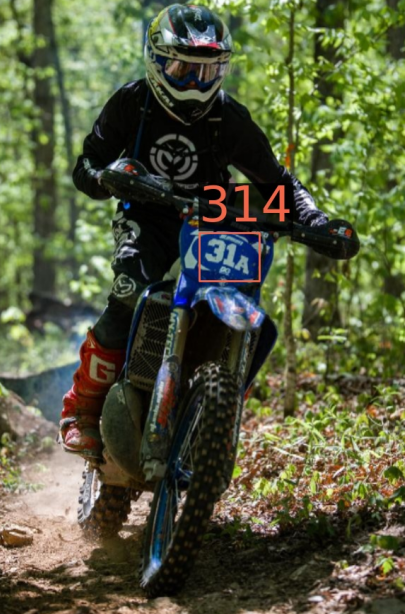}
        \caption{}
        \label{img:wrongletter}
    \end{subfigure}
    \hfill
    \begin{subfigure}[b]{0.115\textwidth}
        \centering
        \includegraphics[width=\textwidth]{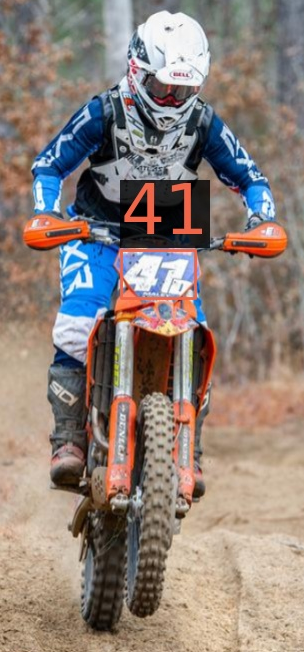}
        \caption{}
        \label{img:missedletter}
    \end{subfigure}
    \hfill
    \begin{subfigure}[b]{0.295\textwidth}
        \centering
        \includegraphics[width=\textwidth]{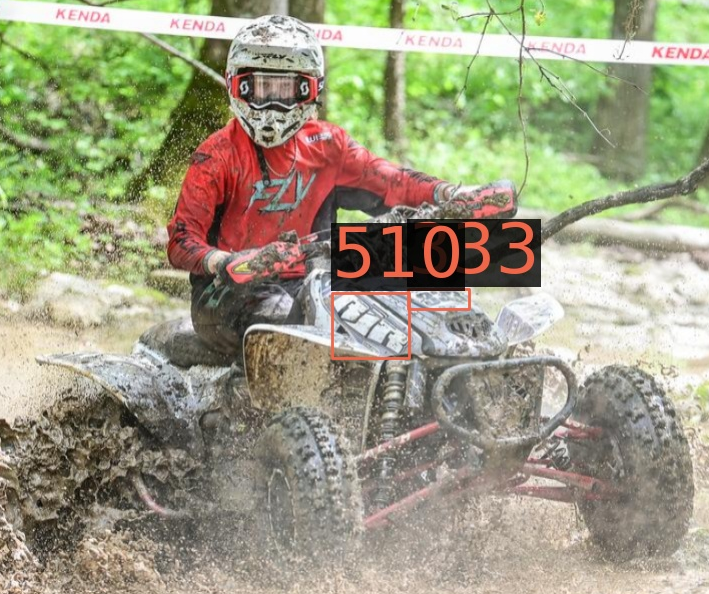}
        \caption{}
        \label{img:quadshard}
    \end{subfigure}
    \caption{Analysis of common non-mud failures: (a) Incorrect side number recognition. (b) Overlapping ``stacked'' numbers confuse the model. (c) A letter is misrecognized as a number. (d) The letter portion of the racer number is missed. (e) Complex graphics on quad confuse model.}  
    \label{fig:cleanfails}
\end{figure*}

\begin{figure}[thb]
    \centering
    \begin{subfigure}[b]{0.235\textwidth}
        \centering
        \includegraphics[width=\textwidth]{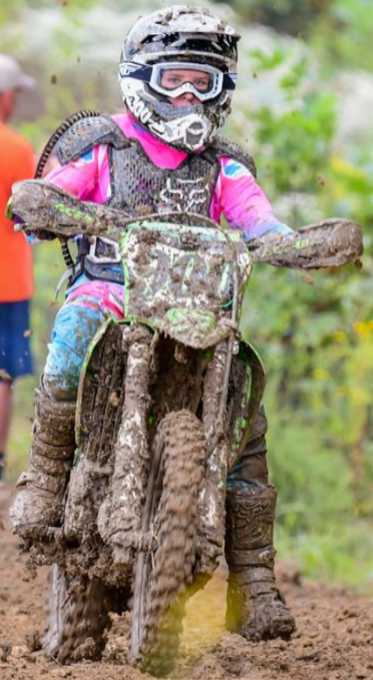}
    \end{subfigure}
    \hfill
    \begin{subfigure}[b]{0.235\textwidth}
        \centering
        \includegraphics[width=\textwidth]{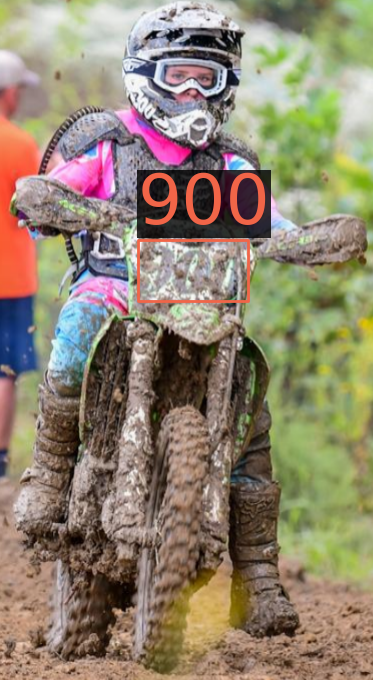}
    \end{subfigure}
    \caption{Example showcasing the fine-tuned model learning to see through the mud. The left image depicts the predictions from the off-the-shelf YAMTS model before fine-tuning, which does not recognize any text. The right image displays results from the fine-tuned YAMTS model, which can see through the heavy mud occlusion and properly detect and recognize the racer number. This demonstrates improved robustness to real-world mud occlusion after domain-specific fine-tuning.}
    \label{fig:see-through-mud}
\end{figure}

\begin{figure}[thb]
    \centering
    \begin{subfigure}[b]{0.5\textwidth}
        \centering
        \includegraphics[width=\textwidth]{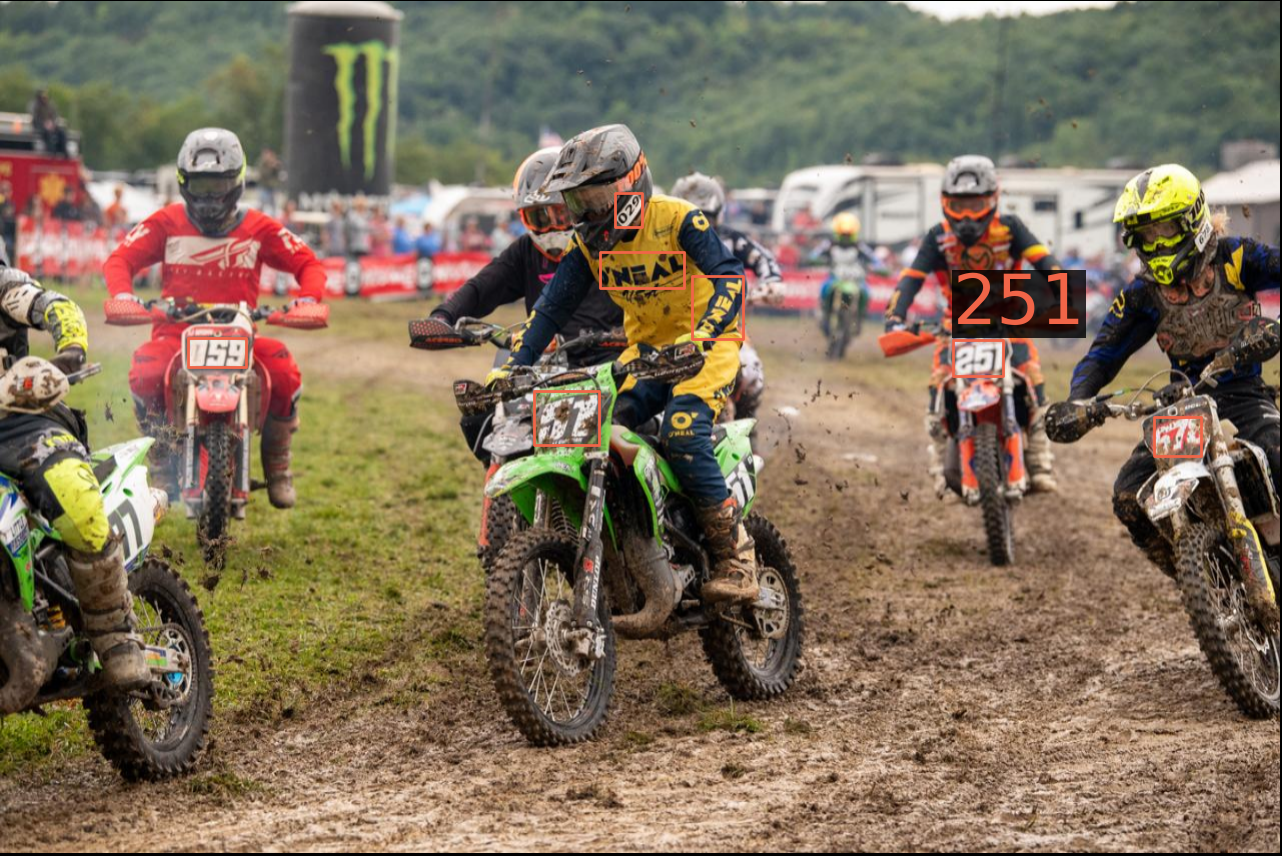}
    \end{subfigure}
    \hfill  
    \begin{subfigure}[b]{0.5\textwidth}
        \centering
        \includegraphics[width=\textwidth]{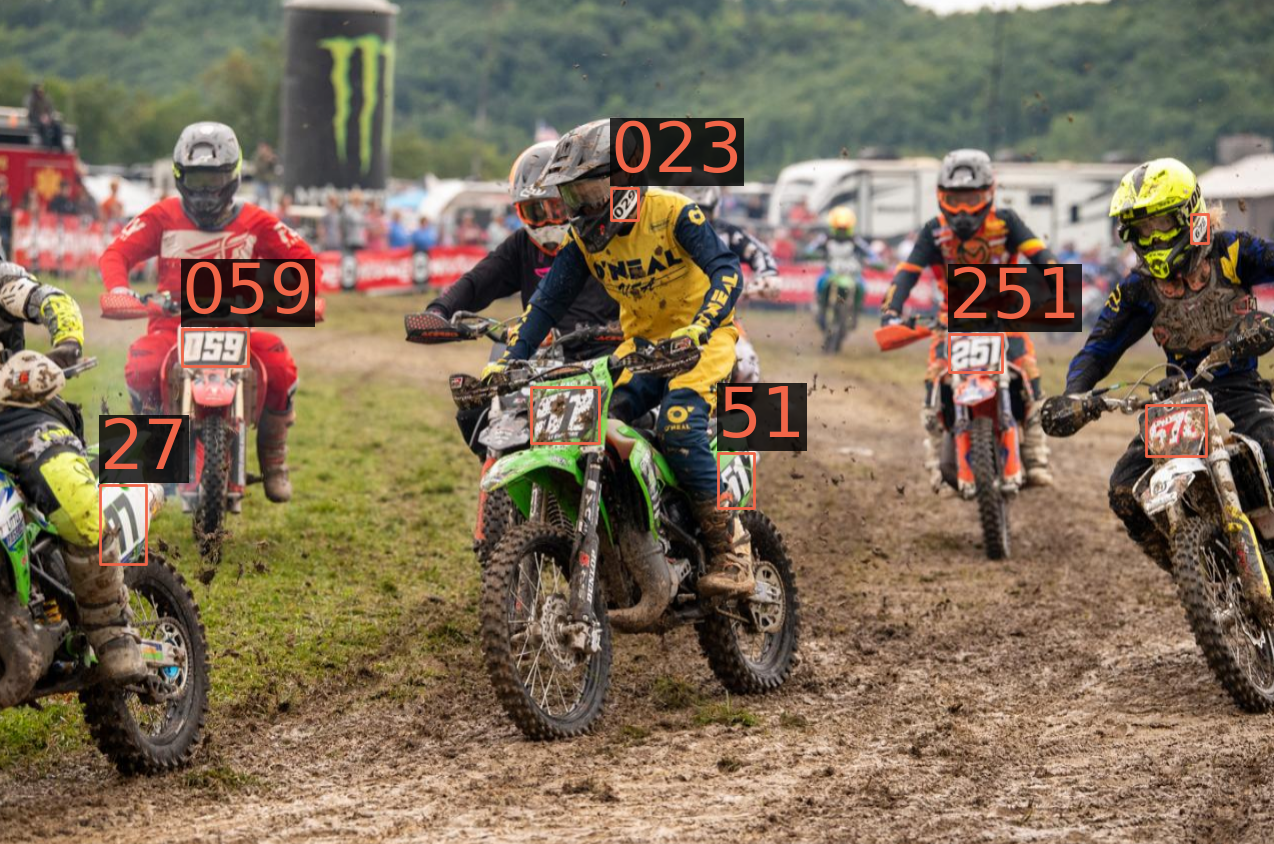}
    \end{subfigure}
    \caption{Example showcasing model successes and failures on a complex muddy image. The top image shows detected text from the off-the-shelf YAMTS model before fine-tuning, which recognizes only 1 number correctly (``251"). The bottom image displays results from the fine-tuned YAMTS model, which detects all 8 visible numbers but only correctly recognizes 3 of them. This highlights the benefits of domain-specific fine-tuning, as the pre-trained model struggles. However, even the fine-tuned model has difficulty accurately recognizing highly degraded text, exposing substantial room for improvement.}
    \label{fig:muddy-start}
\end{figure}

Figure~\ref{fig:mudfails} showcases common mud-related successes and failures. 
In some cases, 
the fine-tuned models can see through mud occlusions to properly recognize 
the racer number, as shown in Figure~\ref{img:rtm2}.
However, mud often prevents smaller helmet numbers 
from being recognized (Fig~\ref{img:rtm1}, \ref{img:mudfail}). 
Odd orientations also confuse models (Fig~\ref{img:mudslidefail}). 
Overall, heavy mud occlusion remains the biggest challenge.
Figure~\ref{fig:cleanfails} reveals other common failures 
like missing side numbers (Fig~\ref{img:backbikefail}), 
overlapping numbers (Fig~\ref{img:stackednums}), 
confusion between letters and numbers (Fig~\ref{img:wrongletter}), 
missing letter portions (Fig~\ref{img:missedletter}), 
and distractions from graphics (Fig~\ref{img:quadshard}).
In summary, 
the analysis reveals promising capabilities 
but also exposes key areas for improvement, 
particularly among extreme mud and small text.
Substantial opportunities remain to enhance OCR 
for this challenging real-world application.

\section{Additional Reid results}
\label{sec:additional-reid-results}

\begin{figure*}[thb]
    \centering
    \includegraphics[width=\textwidth]{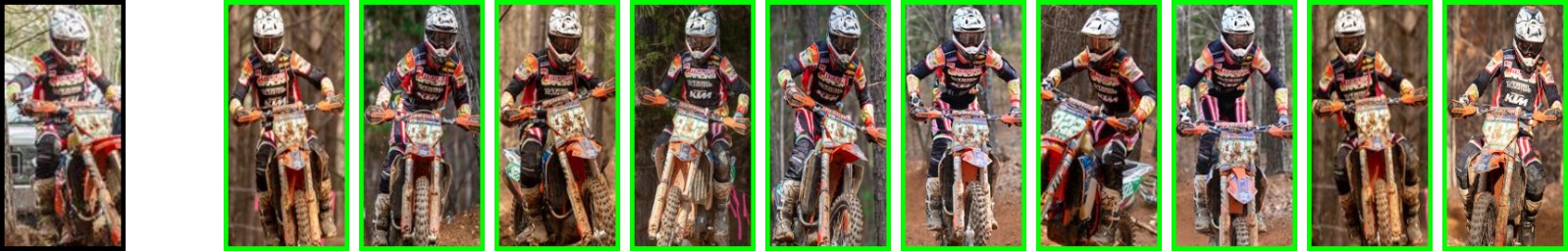}
    \caption{Example of successful re-id by the fine-tuned model under moderate mud occlusion. The 10 top retrievals correctly identify the query rider despite the mud, pose, and other variations. Green boundaries signify correct matches and red incorrect.}
    \label{fig:lightmud}
\end{figure*}

\begin{figure*}[thb]
    \centering
    \includegraphics[width=\textwidth]{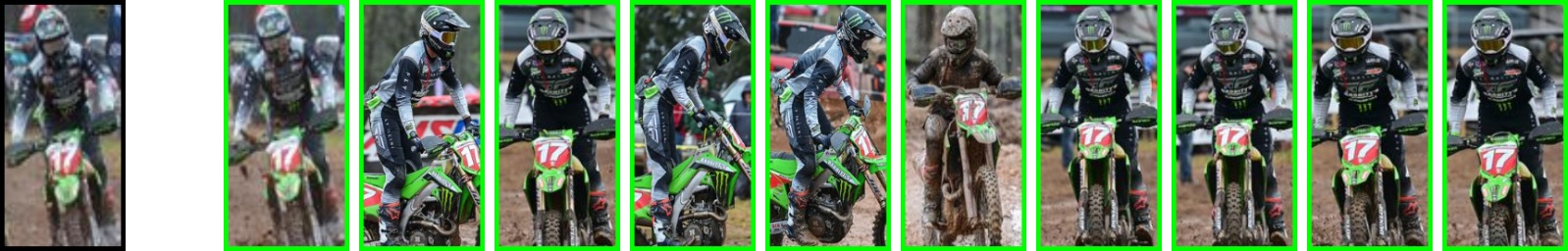}
    \caption{Example of the model correctly matching a clean image of a rider to a muddy image of the same rider when the pose is similar between the query and gallery image. Green boundaries signify correct matches and red incorrect.}
    \label{fig:clean-to-muddy}
\end{figure*}

\begin{figure*}[thb]
    \centering
    \includegraphics[width=\textwidth]{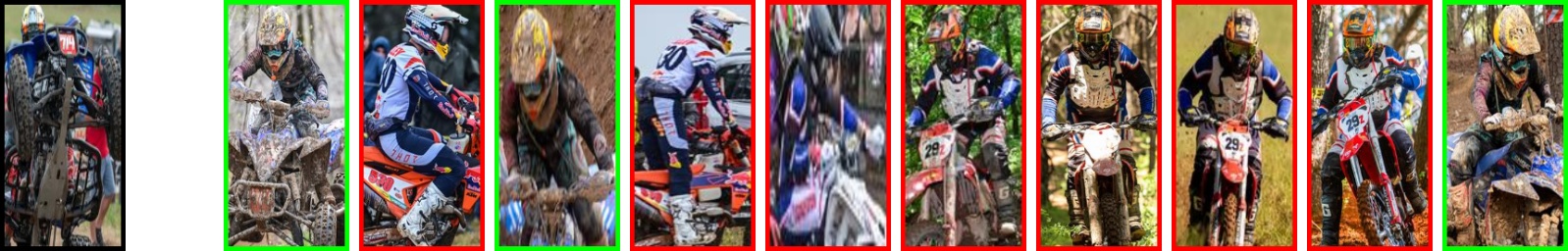}
    \caption{Example of a failure case due to extreme pose variation in the query image. The rider is captured doing a wheelie, leading to incorrect matches despite no mud occlusion. Green boundaries signify correct matches and red incorrect.}
    \label{fig:wheelie-pose}
\end{figure*}

\begin{figure*}[thb]
    \centering
    \includegraphics[width=\textwidth]{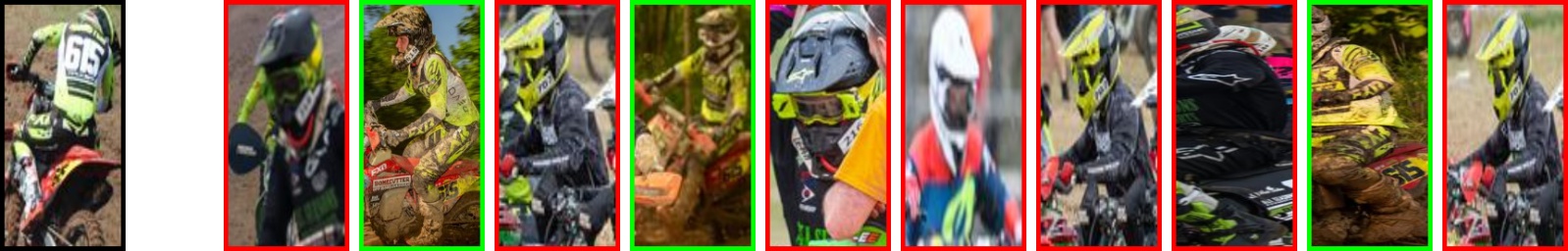}
    \caption{Failure case due to pose variation between the query and gallery images. The backward-facing query rider is not matched to forward-facing images of the same identity. Green boundaries signify correct matches and red incorrect.}
    \label{fig:poses-are-hard}
\end{figure*}

\begin{figure*}[thb]
    \centering
    \includegraphics[width=\textwidth]{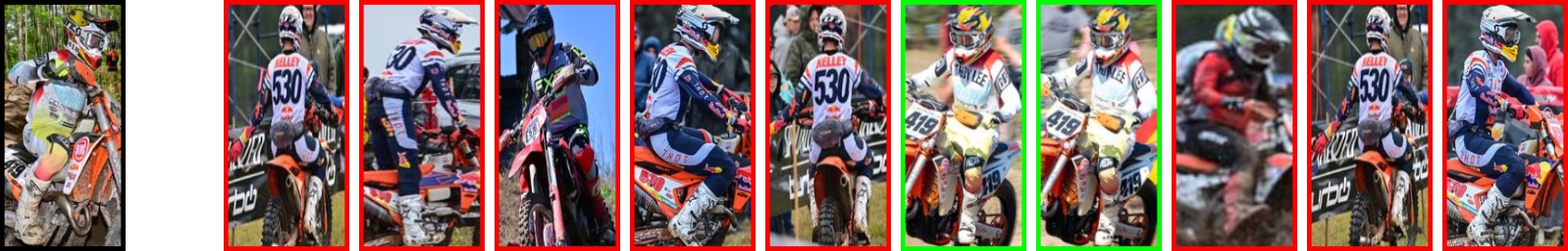}
    \caption{Example failure case due to two different riders having very similar jerseys and gear, leading to confusion between their identities. Green boundaries signify correct matches and red incorrect.}
    \label{fig:similar_outfits}
\end{figure*}

\begin{figure*}[thb]
    \centering
    \includegraphics[width=\textwidth]{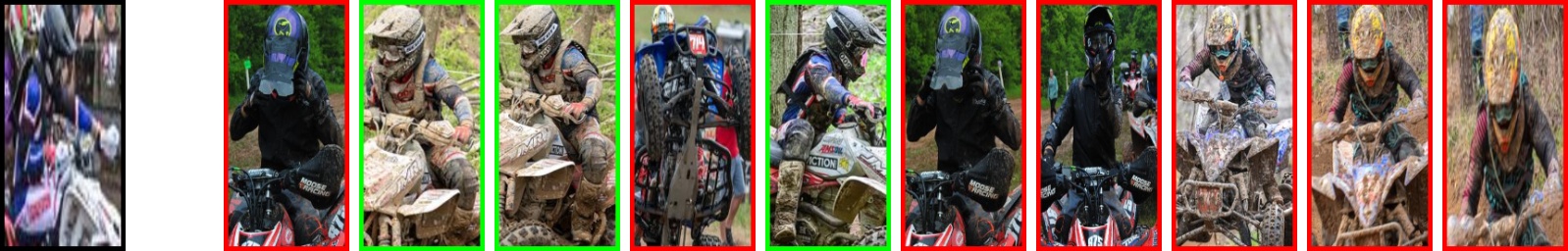}
    \caption{Failure case due to low resolution of the query image preventing distinguishing details from being visible. The small, distant crop of the rider cannot be matched accurately. Green boundaries signify correct matches and red incorrect.}
    \label{fig:lowres}
\end{figure*}

As seen in Figure \ref{fig:wheelie-pose}, 
a rider doing a wheelie is not matched 
to more standard riding poses. 
Even common pose differences like 
front versus back views 
are challenging (Figure \ref{fig:poses-are-hard}).
Images with small, 
distant crops of riders lack fine details 
for discrimination. 
Figure \ref{fig:lowres} 
shows a failure case where the query is low resolution.
In some cases, 
different riders with very similar 
gear is confused. 
This is common as racers supported by the same 
team will typically purposefully 
coordinate their appearance. 
An example is shown in 
Figure \ref{fig:similar_outfits}.

\end{document}